%% file: main_pure.tex
\documentclass[lettersize,journal]{IEEEtran}
\usepackage{amsmath,amsfonts}
\usepackage{algorithmic}
\usepackage{array}
\usepackage[caption=false,font=normalsize,labelfont=sf,textfont=sf]{subfig}
\usepackage{textcomp}
\usepackage{stfloats}
\usepackage{url}
\usepackage{verbatim}
\usepackage{graphicx}
\usepackage{color}
\usepackage{multirow}
\usepackage{array}
\usepackage{bbding}
\usepackage{amssymb}

\newcommand{\data}{road network}
\newcommand{\fulldata}{road network}
\newcommand{\Fulldata}{Road network}
\newcommand{\seqdata}{RoadNet Sequence}
\newcommand{\sarseqdata}{Semi-Autoregressive RoadNet Sequence}
\newcommand{\name}{RoadNetTransformer}

\newcommand{\abb}{RNTR}
\newcommand{\arabb}{AR-RNTR}
\newcommand{\sarabb}{SAR-RNTR}
\newcommand{\narabb}{NAR-RNTR}
\newcommand{\adjf}{lossless}
\newcommand{\sdseqdata}{SD-Map Sequence}
\newcommand{\strategy}{Topology-Inherited Training}
\newcommand{\stabb}{TIT}
\newcommand{\toponame}{SDMap-RoadNetTransformer}
\newcommand{\topoabb}{SD-RNTR}

\def\eg{\textit{e.g.}}
\def\ie{\textit{i.e.}}

\usepackage{colortbl}
\definecolor{maroon}{cmyk}{0,0.87,0.68,0.32}

\newcommand{\revision}[1]{\textcolor{black}{#1}}
\newcommand{\ext}[1]{\textcolor{black}{#1}}

\newcommand{\PreserveBackslash}[1]{\let\temp=\\#1\let\\=\temp}
\newcolumntype{C}[1]{>{\PreserveBackslash\centering}p{#1}}

\def\BibTeX{{\rm B\kern-.05em{\sc i\kern-.025em b}\kern-.08em
    T\kern-.1667em\lower.7ex\hbox{E}\kern-.125emX}}
\usepackage{balance}
\begin{document}
\title{Translating Images to Road Network: \\A Sequence-to-Sequence Perspective}
\author{Jiachen Lu \quad Ming Nie \quad Bozhou Zhang \quad Renyuan Peng \\ Xinyue Cai \quad Hang Xu \quad Feng Wen \quad Wei Zhang \quad Li Zhang
\thanks{Li Zhang is the corresponding author (lizhangfd@fudan.edu.cn).\\
Jiachen Lu, Ming Nie, Bozhou Zhang, Reyuan Peng and Li Zhang are with the School of Data Science, Fudan University.\\
Xinyue Cai, Hang Xu, Feng Wen and Wei Zhang are with Huawei Noah’s Ark Lab.}}

\markboth{Journal of \LaTeX\ Class Files,~Vol.~18, No.~9, September~2020}%
{How to Use the IEEEtran \LaTeX \ Templates}

\maketitle

\input{file/0-abstract}
\input{file/1-intro}
\input{file/2-related}
\input{file/3-method}

\input{file/3-new-method}
\input{file/4-exp}
\input{file/5-conclusion}

\bibliographystyle{IEEEtran}
\bibliography{egbib}

\newpage
\input{file/6-appendix-new}

\end{document}

%% file: file/0-abstract.tex
\begin{abstract}
The extraction of \fulldata{} is essential for the generation of high-definition maps since it enables the precise localization of road landmarks and their interconnections.
However, generating \data{} poses a significant challenge due to the conflicting underlying combination of Euclidean (\eg, road landmarks location) and non-Euclidean (\eg, road topological connectivity) structures. 
Existing methods struggle to merge the two types of data domains effectively, but few of them address it properly.
Instead, our work establishes a unified representation of both types of data domain by projecting both Euclidean and non-Euclidean data into an integer series called \seqdata{}.
Further than modeling an auto-regressive sequence-to-sequence Transformer model to understand \seqdata{}, we decouple the dependency of \seqdata{} into a mixture of auto-regressive and non-autoregressive dependency.
Building on this, our proposed non-autoregressive sequence-to-sequence approach  leverages non-autoregressive dependencies while fixing the gap towards auto-regressive dependencies, resulting in success in both efficiency and accuracy.
\ext{
We further identify two main bottlenecks in the current \name{} on a non-overfitting split of the dataset: poor landmark detection limited by the BEV Encoder and error propagation to topology reasoning. Therefore, we propose \strategy{} to inherit better topology knowledge into \name{}. Additionally, we collect SD-Maps from open-source map datasets and use this prior information to significantly improve landmark detection and reachability.
}
Extensive experiments on the nuScenes dataset demonstrate the superiority of \seqdata{} representation and the non-autoregressive approach compared to existing state-of-the-art alternatives.
Our code is publicly available at open-source \textcolor{black}{\url{https://github.com/fudan-zvg/RoadNetworkTRansformer}}.
\end{abstract}


%% file: file/1-intro.tex
\section{Introduction}
\input{figure/intro-fig}
With the rising prevalence of self-driving cars, a deep knowledge of the road structure is indispensable for autonomous vehicle navigation~\cite{cui2019multimodal,hong2019rules,rella2021decoder,chen2020learning}. 
\Fulldata{}~\cite{casas2021mp3,ma2019exploiting, ravi2018real} extraction is required to estimate highly accurate {\em road landmark locations, centerline curve shapes, and road topological connection} for self-driving vehicles. 
However, the ability to understand \data{} in real-time using onboard sensors is highly challenging.

In the literature, \data{} is {\bf\em differentiated} from methods that focus solely on grid-like Euclidean data~\cite{bronstein2017geometric, bronstein2021geometric} such as lane detection~\cite{xu2020curvelane,liu2021condlanenet,xu2022rclane} or BEV semantic understanding~\cite{philion2020lift, saha2022translating,li2022bevformer,lu2022learning}.
Instead, \data{} emphasizes a more comprehensive understanding of both {\bf\em Euclidean domain} and {\bf\em non-Euclidean domain}.
As shown in Figure~\ref{fig:intro}, accurate road landmark locations such as crossroads, stop-lines, fork-roads, and the shape of centerline curves pertain to the Euclidean domain, whereas road topology belongs to the non-Euclidean domain.
In mathematical terms~\cite{bronstein2021geometric, bronstein2017geometric}, {\em Euclidean data} is defined in $\mathbb{R}^2$. On the other hand, {\em non-Euclidean data}, such as graphs that indicate connectivity among nodes, will lose crucial information if projected to $\mathbb{R}^n$, such as edge curve information.

Unfortunately, existing attempts~\cite{can2021structured, can2022topology} to extract \data{} using onboard sensors have not been able to achieve a harmonious integration between Euclidean and non-Euclidean data.
STSU~\cite{can2021structured} divides the \data{} construction into two stages: center-lines detection and center-line connectivity reasoning—which ignores the cooperation between Euclidean and non-Euclidean domains.
TPLR~\cite{can2022topology} uses a Transformer or Polygon-RNN~\cite{acuna2018efficient} to combine topology reasoning with center-line localization, but the embedded conflicts between Euclidean data representations and connectivity representations undermine the model performance.

In this work, we propose that the dilemma in existing
works arises due to the absence of a unified representation
of both Euclidean and non-Euclidean data. 
Instead, we introduce a Euclidean-nonEuclidean unified representation with merits of losslessness, efficiency and interaction.
The unified representation, named as \seqdata{}, projects both Euclidean and non-Euclidean aspects of road network to integer series domain $\mathbb{Z}^n$.
The ``losslessness" aspect is ensured by establishing a {\bf\em bijection} from road network to \seqdata{}.
``Efficiency" is achieved by limiting \seqdata{} length to the shortest $\mathcal{O}(|\mathcal{E}|)$ (where $\mathcal{E}$ is the set of all centerlines), through a specially designed topological sorting rule.
``Interaction" reveals the interdependence between Euclidean and non-Euclidean information within a single sequence.

Based on the auto-regressive dependency of topological sorting, we leverage the sequence-to-sequence generation power of Transformer~\cite{vaswani2017attention, brown2020language, chen2021pix2seq} to understand \seqdata{} from onboard round-view cameras, called {\bf\em R}oad{\bf\em N}etwork{\bf\em TR}ansformer ({\bf\em \abb{}}).

\ext{
However, the varying order of node and edge generation results in different auto-regressive dependencies in sequence-to-sequence generation. 
In practice, we examine two types of dependencies: coupled \seqdata{} and decoupled \seqdata{}. 
The coupled \seqdata{} generates each node and immediately outputs the edge connected to it, whereas the decoupled \seqdata{} first generates all nodes and then outputs the edges connected to each node.
}

\ext{
Moreover, the decoupled \seqdata{} can be upgrated to accelerate the slow auto-regressive process.}
Our observation is that the dependency of \data{} can also be {\em decoupled} into a semi-autoregressive format, that retains auto-regressive functionality within local contexts while simultaneously conducting multiple generations in parallel.
This semi-autoregressive model is called Semi-Autoregressive \name{} ({\bf\em \sarabb{}}).
This approach not only accelerates the inference speed by 6 times, but it also significantly {\em boosts the accuracy} to a new level based on the better dependency modeling.
Going beyond \sarabb{}, we employ a masked training technique on \sarabb{} to mimic the remaining auto-regressive dependency through iterative prediction~\cite{lee2018deterministic}. 
This gives rise to our Non-Autoregressive \name{} ({\bf\em \narabb{}}) model, which achieves real-time inference speed ($47\times$ faster) while maintaining the high performance.


To evaluate \data{} extraction quality, apart from inheriting the former metrics~\cite{can2021structured} based on lane detection, 
we instead design a family of metrics directly based on the definition of the road network -- Landmark Precision-Recall and Reachability-Precision-Recall -- to evaluate (i) road landmarks localization accuracy, and (ii) path accuracy between any reachable landmarks.

We make the following {\bf contributions}: (i) we introduce  \seqdata{}, a \adjf{}, efficient and unified representation of both Euclidean and non-Euclidean information from \data{}.
(ii) We propose a Transformer-based \name{} which can decode \seqdata{} from multiple onboard cameras.
(iii) By decoupling auto-regressive dependency of \seqdata{}, our proposed Non-autoregressive \name{} accelerates the inference speed to real-time while boosting the accuracy with a significant step from the auto-regressive model.
(iv) Extensive experiments on nuScenes~\cite{caesar2020nuscenes} dataset validate the superiority of \seqdata{} representation and \name{} over the alternative methods with a considerable margin. 

\ext{
In the extension, we address the major bottlenecks of sequence-to-sequence models for road network extraction. 
During our exploration, we split the nuScenes~\cite{caesar2020nuscenes} dataset into non-overlapping scenes to evaluate the non-overfitting performance.
Unlike before~\cite{roddick2020predicting}, we also ensure that different daytime and weather conditions are equally distributed in the training and evaluation sets.
The bottlenecks arise from poor road perception, and the errors subsequently propagate to the understanding of the topology.
To mitigate error propagation in topology learning, we propose a new training strategy called {\bf\em \strategy{}} ({\bf\em \stabb{}}), which inherits topology knowledge from a large dataset trained model.
\stabb{} first trains a high-performance {\bf\em\toponame{}} ({\bf\em \topoabb{}}) by using bird's-eye-view (BEV) maps as input instead of round-view images to simplify Euclidean data perception. 
Additionally, we use the \texttt{sweep} data in nuScenes~\cite{caesar2020nuscenes}, which is nine times larger than the traditional nuScenes dataset, to learn better non-Euclidean perception.
We then distill the BEV representation from the encoder pretrained with lidar maps to the traditional round-view image input BEV Encoder. 
Finally, we assemble the BEV Encoder with round-view image input and \name{}, boosting performance across different modality inputs.
Apart from enhancing non-Euclidean data learning, the distillation of round-view image BEV representation learning can also improve the performance of Euclidean data.
}

\input{figure/sd_intro_eg}

\ext{
Although the \stabb{} partly solve the bottlenecks, the further use of standard definition maps (SD-Maps) from daily navigation applications like Google Maps provides essential prior information about basic landmarks and topology. 
As shown in Figure~\ref{fig:intro}, unlike \data{}, which contains lane-level and centimeter-level accuracy information, SD-Maps offer only road-level information, indicating road connections with meter-level accuracy. Consequently, SD-Maps cannot provide accurate lane numbers or precise landmark locations.
We collect the corresponding SD-Map data from OpenStreetMap~\cite{openstreetmap} (similar to Google Map) and manually align it with the nuScenes~\cite{caesar2020nuscenes} dataset.
By treating SD-Maps as a prompt learning problem and integrating them into the Transformer model, we achieve significant performance improvements without altering the model structure or requiring complex data processing.}

\ext{
In conclusion, we make the following extension: 
\textbf{(i)}
We examine the advantages of decoupled \seqdata{};
\textbf{(ii)}
We propose a brand-new training strategy, \strategy{}, to mitigate error propagation in topology learning, enhancing both Euclidean and non-Euclidean data understanding;
\textbf{(iii)}
We introduce \toponame{}, which realizes the inheritance of topology knowledge across modalities;
\textbf{(iv)}
We collect SD-Map data from OpenStreetMap~\cite{openstreetmap} and treat SD-Maps as a prompt learning problem without altering the model structure or complex data processing.
}

%% file: figure/intro-fig.tex
\begin{figure}[htb]
    \centering
    \includegraphics[width=\linewidth]{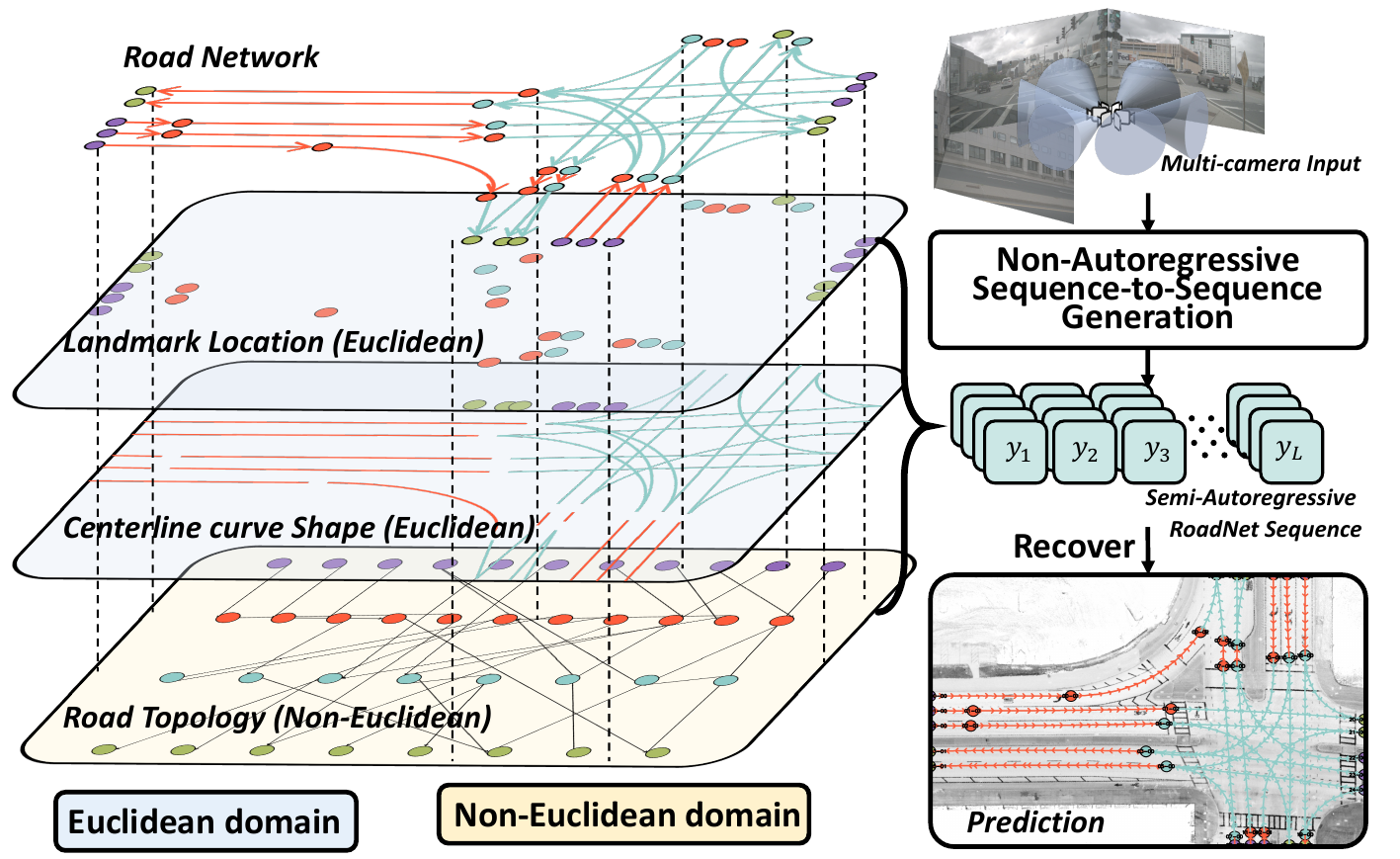}
    \caption{High-definition Road Network Topology contains {\bf\em Euclidean data}: locations of landmarks and shapes of curves and {\bf\em non-Euclidean data}: road topology. 
    A special sequence, \seqdata{}, is proposed as a unified representation of both domains.
    Then we use a Non-Autoregressive sequence-to-sequence approach to extract \seqdata{} from multi-camera input efficiently and accurately.
    }
    \label{fig:intro}
    \vspace{-6mm}
\end{figure}

%% file: figure/sd_intro_eg.tex
\begin{figure}[t]
    \centering
    \includegraphics[width=\linewidth]{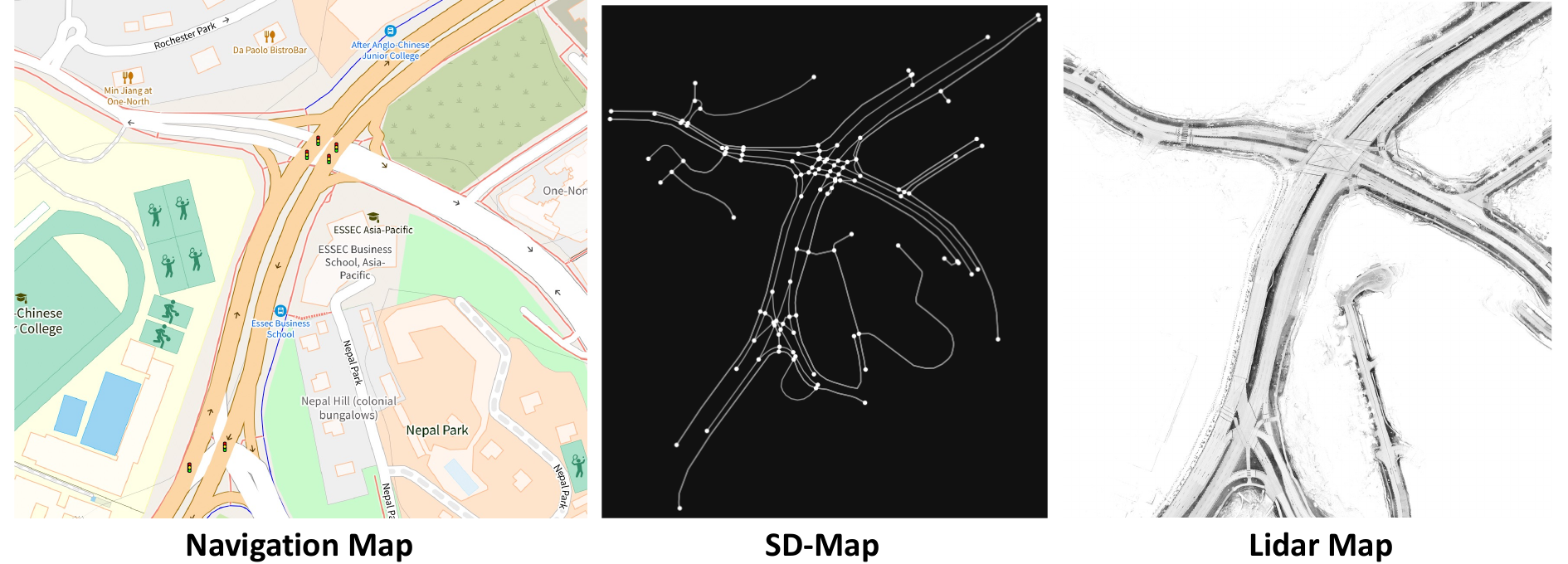}
    \caption{
    \ext{
Comparison between Navigation Map (\eg, Google Maps~\cite{googlemaps} or OpenStreetMap~\cite{openstreetmap}), SD-Map (extracted from OpenStreetMap), and Lidar Map (collected by the nuScenes~\cite{caesar2020nuscenes} dataset).}
    }
    \label{fig:sd_intro_eg}
    \vspace{-6mm}
\end{figure}

%% file: file/2-related.tex
\section{Related work}
\noindent \textbf{Vision-based ego-car (BEV) feature learning.}
A rising tendency is conducting self-driving downstream tasks \cite{roddick2018orthographic, li2022bevformer, lu2022learning, huang2021bevdet, philion2020lift, saha2022translating, hu2021fiery, zhou2023suit, gao2020vectornet, liao2022maptr} under ego-car coordinate frame.
To learn ego-car feature learning from onboard cameras, ~\cite{wang2019pseudo, reading2021categorical} project image features to ego-car coordinate based on depth estimation. OFT~\cite{roddick2018orthographic}, LSS~\cite{philion2020lift} and FIERY~\cite{hu2021fiery} predict depth distribution to generate the intermediate 3D representations. 
\cite{saha2022translating, roddick2020predicting, li2022bevformer, lu2022learning} resort neural network like Transformer to learn ego-car feature without depth.
In order to learn the topology of the road network under the ego-car coordinate frame, we use the straightforward method of applying LSS~\cite{philion2020lift, huang2021bevdet} to extract ego-car features from the multiple onboard cameras.

\noindent \textbf{\Fulldata{} extraction.}
Researchers have explored the utilization of DNNs to decode and recover maps from aerial images and GPS trajectories~\cite{ruan2020learning, wu2020deepdualmapper}.
Moreover, STSU~\cite{can2021structured} first detects centerline from front-view image with Transformer and then predicts the association between centerline with MLP layer, followed by a final merge to estimate \data{}.
Based on STSU, TPLR~\cite{can2022topology} introduces minimal cycle to eliminate ambiguity in its connectivity representation.
Existing methods spend great effort to deal with the problem in the non-Euclidean domain but ignore the cooperation between Euclidean and non-Euclidean.
Previous approaches, such as STSU~\cite{can2021structured} and TPLR~\cite{can2022topology}, focus on solving the problem within the non-Euclidean domain, typically handling road network connectivity through centerline detection or minimal cycles. While these methods tackle topological reasoning and road connectivity, they often neglect the interdependencies between Euclidean and non-Euclidean information. For example, STSU detects centerlines using Transformer-based models and then predicts the association between centerlines through an MLP, but it does not effectively merge road geometry with topological structure. TPLR addresses connectivity ambiguities by introducing minimal cycles, but it does not fully integrate Euclidean data with non-Euclidean topological data.
In contrast, our work proposes a unified representation that integrates both Euclidean and non-Euclidean road network information into a single framework, overcoming the limitations of previous methods. Instead of treating these domains separately, as done by STSU~\cite{can2021structured} and TPLR~\cite{can2022topology}, we introduce a new method that projects both Euclidean and non-Euclidean data into an integer series called \seqdata{}. By doing so, we create a unified, lossless, and interactive representation that allows for more accurate modeling of road networks.
Further, we introduce a sequence-to-sequence Transformer model that decouples the dependency of \seqdata{} into a mixture of auto-regressive and non-autoregressive dependencies. This innovative approach allows our model to leverage the benefits of both dependencies, ensuring high efficiency and accuracy in predicting road network properties. Our proposed non-autoregressive approach significantly improves the efficiency of training and inference while maintaining the model's predictive accuracy.

\noindent \textbf{Non-Autoregressvie generation.}
Non-Autoregressive generation for Neural Machine Translation (NMT)~\cite{gu2017non} has been proposed as a solution to speed up the one-by-one sequential generation process of autoregressive models.
\cite{kim2016sequence, zhou2019understanding, ren2020study} utilized knowledge distillation to help the NAT model capture target sequence dependency.
\cite{lee2018deterministic, stern2019insertion, gu2019levenshtein,ghazvininejad2019mask} proposed iterative models that refine the output from the previous iteration or a noised target in a step-wise manner.
\cite{wang2018semi} kept the auto-regressive approach for global modeling, but introduced the parallel output of a few successive words at each time step.
\cite{shu2020latent, bao2022glat, ran2021guiding, bao2021non} used latent variables as intermediates to reduce the dependency on the target sequence.
The current state of non-autoregressive NMT models is far from perfect, with a significant gap in performance compared to their auto-regressive counterparts~\cite{xiao2022survey}.
But in the case of our \seqdata{}, the auto-regressive dependency can be decoupled, leading to both acceleration and improved performance.

%% file: file/3-method.tex
\section{Method}
\input{figure/AR-RTSeq}
\input{figure/SAR-RTTR-arc}
In this section, we will introduce: (i) mathematics modeling of \data{}, (ii) {\bf\em \seqdata{}}, (iii) architecture of {\bf\em \name{}}, (iv) metrics.

\subsection{Mathematics Modeling of Road Network}
The \data{} comprises road landmarks (\eg, crossroad, stop-lines, fork-point, merge-point) and centerlines connecting among them~\cite{can2021structured, can2022topology}.
Given that traffic on each lane is always one-way, the graph can be formulated as a Directed Acylic Graph (DAG), \ie, $\mathcal{G}=(\mathcal{V}, \mathcal{E})$ where vertex set $\mathcal{V}$ is the set of all road landmarks and edge set $\mathcal{E}$ is the set of all centerlines.
Each vertex $v=(v_x, v_y, v_c)\in\mathcal{V}$ contains two properties: (i) location, \ie, $(v_x\in\mathbb{R}, v_y\in\mathbb{R})$, (ii) category, \ie, $v_c$.
As shown in Figure~\ref{fig:ar-rtseq}, we divide vertices into 4 categories which will be introduced later.
Each edge $e=(e_s, e_t, e_{px}, e_{py})\in\mathcal{E}$ contains two properties: (i) source $e_s\in\mathcal{V}$ and target $e_t\in\mathcal{V}$ vertex of the edge, (ii) Bezier middle control point $(e_{px}\in\mathbb{R}, e_{py}\in\mathbb{R})$~\cite{can2021structured} controlling the shape of edge curve.
By and large, the {\bf\em Euclidean data} of $\mathcal{G}$ contains $v_x, v_y, e_{px}, e_{py}$ while the {\bf\em non-Euclidean data} contains $v_c, e_s, e_t$.

\input{figure/DAR-RTSeq}

\subsection{Coupled \seqdata{}}
In this section, we present {\bf\em \seqdata{}} to integrate Euclidean and Non-Euclidean data into a unified representation.
The projection from \data{} to coupled \seqdata{} includes: (i) from DAG to Directed Forest, (ii) topological sorting of Directed Forest and (iii) sequence construction.

\noindent \textbf{From DAG to Directed Forest.}
A typical DAG always has the ambiguity relationship between vertices and edges ($|\mathcal{V}| \nleftrightarrow |\mathcal{E}|$) so that preserving all edges and vertices leads to redundancy due to the repeated use of vertices when indicating edge direction.
In comparison, as shown in Figure~\ref{fig:ar-rtseq}, the Tree structure has a clear relationship between them ($|\mathcal{V}| \leftrightarrow |\mathcal{E}|$) which avoids this redundancy.
Hence, we innovatively transform DAG to a Directed Forest $\mathcal{G}_f=(\mathcal{V}_f, \mathcal{E}_f)$ (a collection of disconnected directed trees).
This transformation has the benefit of enabling us to find a {\bf\em bijection} between $\mathcal{E}_f$ and $\mathcal{V}_f\backslash\{v\in\mathcal{V}_f\ |\ id(v)=0\}$ (set of all vertices which are not root, where $id(v)$ means incoming degree of the vertex).
The bijection $f: \mathcal{V}\rightarrow \mathcal{E}$ can be $f(v)=(Parent(v), v)$ for any vertices $v\in\mathcal{V}_f\backslash\{v\in\mathcal{V}_f\ |\ id(v)=0\}$ where $Parent(v)$ is its unique parent and $f^{-1}(e)=v_1$ for any $e=(v_0, v_1)\in\mathcal{E}$.

As shown in the top of Figure~\ref{fig:ar-rtseq}, the transformation happens for all the merge-points on the road, \ie, $\{v\in\mathcal{V}\ |\ id(v) > 1\}$ whose parents are non-unique.
We replicate their parents (except the first parent) to their children and delete the corresponding edges. 
Then we assign these created vertices with a specific label \texttt{Clone} so that the direction of the edge between the merge-points and themselves will be reverted during recovery.

\noindent \textbf{Topological sorting of Directed Forest.}
We use Depth-first search (DFS) to obtain the sequential order of vertices $\mathcal{V}_f$ in the Directed Forest, \ie, $[v_1, v_2, \cdots, v_n]$.
But the \texttt{Clone} point labeled in the last paragraph will not be traversed.
Instead, as shown in the bottom of Figure~\ref{fig:ar-rtseq}, \texttt{Clone} point will be traversed following its original point.

Thanks to the bijection $f$ between $\mathcal{E}_f$ and $\{v\in\mathcal{V}_f\ |\ id(v)=1\}$, we can build pairs for all edges and vertices, \ie, $(v, f(v))$ for $v\in\{v\in\mathcal{V}_f\ |\ id(v)=1\}$ and $(v, \text{None})$ for $v\in\{v\in\mathcal{V}_f\ |\ id(v)=0\}$.
These pairs construct the \data{} with no redundancy.

\noindent\textbf{Non-unique sorting.}
In the application of Depth-First Search (DFS), the sequence in which points under the same parent are searched first can result in non-unique sorting. 
We investigate two strategies to address this: a random ordering strategy and an ordering based on coordinates, specifically prioritizing points closest to the Bird's Eye View (BEV) front right.

\noindent \textbf{Sequence construction.}
We then use 6 integers to represent each vertex-edge pair.
6 integers are made up with: (i) two integers for location of vertex, \ie, \texttt{int}$(v_x)$, \texttt{int}$(v_y)$;
(ii) one integer for category $v_c$ of of vertex and we set categories as \texttt{Ancestor}, \texttt{Lineal}, \texttt{Offshoot}, \texttt{Clone};
(iii) one integer for the index of parent (None for root), $v_d=$\texttt{Index}$(Parent(v))$ where index is its topological order (the \texttt{Clone} vertex shares the same index as origin to indicate that they are identical);
(iv) two integers for coefficient of Bezier curve of $e=f(v)$, \ie \texttt{int}$(e_{px})$, \texttt{int}$(e_{py})$.
But, simply discretizing $e_{px}$ and $e_{py}$ can be challenging since the Bezier control points may exceed the Bird's Eye View (BEV) range, and their values may become negative.
As a solution, we discretize $e_{px}$ and $e_{py}$ by applying the \texttt{int} function to $(e_{px}+10)$ and $(e_{py}+10)$, respectively, to avoid negative values.

There exist 4 cases to determine the category $v_c$: 
(i) if $v_i$ is the root of the tree, the category $v_c$ is set as \texttt{Ancestor}, the index of its parent $v_d$ is set as None, its coefficient is ignored.
(ii) if $v_i$ is the first child of its parent, the category $v_c$ is set as \texttt{Lineal}, its $v_d$ is ignored for its parent is exact $v_{i-1}$.
(iii) if $v_i$ is not the first child of its parent, the category $v_c$ is set as \texttt{Offshoot}, its $v_d$ is its parent's index.
(iv) if $v_i$ is the cloned child, the category is set as \texttt{Clone}, its $v_d$ is its original child.
Integer representation of $v_c$ is \texttt{Ancestor}: 0, \texttt{Lineal}: 1, \texttt{Offshoot}: 2, \texttt{Clone}: 3.

\noindent \textbf{Details of sequence construction.}
The discretization of $v_x, v_y$ is simply truncating the integer part.
Integer representation of $v_c$ is \texttt{Ancestor}: 0, \texttt{Lineal}: 1, \texttt{Offshoot}: 2, \texttt{Clone}: 3.
Discretizing $e_{px}$ and $e_{py}$ can be challenging since the Bezier control points may exceed the Bird's Eye View (BEV) range, and their values may become negative.
As a solution, we discretize $e_{px}$ and $e_{py}$ by applying the \texttt{int} function to $(e_{px}+10)$ and $(e_{py}+10)$, respectively, to avoid negative values.

\noindent \textbf{Analysis.}
The proposed \seqdata{} possess the merits of losslessness, efficiency and interaction.
The losslessness of \seqdata{} is guaranteed by establishing a bijection between edges and vertices (excluding the root) of trees. 
Our \seqdata{} has a complexity of $\mathcal{O}(|\mathcal{V}_f|)=\mathcal{O}(|\mathcal{E}|)$, making it the most efficient. 
\revision{
A mixture of Euclidean and non-Euclidean data within a local 6-integer clause facilitates full interaction between the Euclidean and non-Euclidean domains. 
}

The transformation from a Directed Acyclic Graph (DAG) to a Directed Forest shares similarities with the lookahead method in Transformers, especially in capturing sequential dependencies and enhancing future predictions. However, the key difference lies in the methodology: while the lookahead method in Transformers uses dynamic attention to model dependencies, our DAG-to-Directed Forest transformation simplifies the graph structure to reduce redundancy, making the prediction process more efficient. Additionally, while Transformers rely on embeddings and attention, our method uses a predefined structural transformation tailored for road network sequence prediction.

\seqdata{} also possess the auto-regressive dependency.
Since Depth-First search of Trees is always topological sorting, vertices only depend on the previous generated vertices.
Also, our 6 integers come in the order of vertices location, vertices category, index and curve coefficient also preserving the auto-regressive assumption.

\noindent \textbf{Sequence embedding.}
Each vertex-edge pair is represented by 6 integers.
To prevent embedding conflicts between the 6 integers, we divide them into separate ranges.
As a default, we set the embedding size to $576$, which is sufficient to accommodate all the integer ranges.

\subsection{Decoupled \seqdata{}}
\noindent \textbf{Sequence construction.}
\ext{
As shown in Figure~\ref{fig:dec_ar-rtseq}, the sequence of a decoupled \seqdata{} comprises two components: the \textit{\textbf{vertex sequence}} and the \textit{\textbf{edge sequence}}.
}

\ext{
The vertex sequence consists of vertices coordinates arranged in the topological sorting order as discussed before.
Since lane graphs encompass varying vertex quantities, the resulting sequences will exhibit distinct lengths.  
In order to demarcate the conclusion of a vertex sequence, we introduce an \texttt{<EOV>} (End of Vertex Sequence) token.
}

\ext{
An edge can be represented as a set of the source vertex, the target vertex, and the Bezier middle control point.
For each vertex in DAG, We denote the target vertex using the sequential index of its child nodes within the vertex sequence.
The source vertex of each edge is determined by arranging the edge in the order of its source vertex in the vertex sequence.
}

\ext{
So with three parameters $( [Index(\text{child node}), e_{mx}, e_{my}] )$ and the position of this triple, the curve shape and direction of an edge can be determined.
Due to the variable nature of the out-degree for each vertex in the DAG, the number of edges extended by each vertex may vary. Consequently, we employ a \texttt{<Split>} token to ascertain the end of the edge subsequence corresponding to a given vertex as the parent node.
And at the end of the entire edge sequence, we set the \texttt{<EOE>}(End of Edge Sequence) token to determine the termination.
}

\ext{
Besides, we use \texttt{<Start>} token to indicate the beginning of the whole sequence. 
The \texttt{<N/A>} token is utilized to separately pad the vertex sequence and edge sequence, which guarantees that both of them maintain a consistent length.
}

\noindent \textbf{Analysis}
\ext{
The proposed decoupled \seqdata{} also possess the merits of losslessness, efficiency and interaction.
The losslessness of \seqdata{} is guaranteed by establishing a bijection between edges and vertices (excluding the root) of trees. 
The decoupled \seqdata{} has a complexity of $\mathcal{O}(|\mathcal{V}|) + \mathcal{O}(|\mathcal{E}|) = \mathcal{O}(|\mathcal{E}|)$, making it more efficient. 
A mixture of Euclidean and non-Euclidean data within the sentence also facilitates interaction between the Euclidean and non-Euclidean domains. 
}

\subsection{Input and Target Sequence Construction}
\noindent \textbf{Sequence embedding.}
Each vertex-edge pair is represented by 6 integers.
To prevent embedding conflicts between the 6 integers, we divide them into separate ranges which is shown in Table~\ref{tab:embedding}.
As a default, we set the embedding size to $576$, which is sufficient to accommodate all the integer ranges.
\input{table/embedding}

\noindent \textbf{Synthetic noise objects technique.}
The input sequence of \seqdata{} starts with a \texttt{start} token and the target sequence ends with an \texttt{EOS} token.
The \texttt{EOS} token makes the model know where the sequence terminates, but the experiments have shown that it tends to cause the model to stop predicting early without getting the complete sequence. 
Inspired by \cite{chen2021pix2seq}, we use a similar sequence augmentation technique to alleviate the problem called the {\em synthetic noise objects technique}~\cite{chen2021pix2seq}.
The technique composes of {\em sequence augmentation} and {\em sequence noise padding}.
The sequence augmentation adds noise to the position of landmarks and the coefficient of centerlines in input sequence.
Whereas, sequence noise padding is a padding technique.
For input sequences, we generate synthetic noise vertices and append them at the end of the real vertices sequence.   
Each noise vertex includes random locations($v_x$, $v_y$), category($v_c$), index of parent($v_d$) and Bezier curve coefficient($e_{px}$, $e_{py}$).
As for the target sequence, the \texttt{EOS} token is added to the end of the real vertices sequence. We set the target category($v_c$) of each noise vertex to a specific noise class(different from any of the ground-truth labels), and the remaining components($v_x$, $v_y$, $v_d$, $e_{px}$, $e_{py}$) of the noise vertex to the "n/a" class, whose loss is not calculated in the back-propagation.

However, we only use sequence noise padding as sequence augmentation has been shown to cause a decrease in performance.
The introduced modifications of the synthetic noise objects technique are illustrated in Figure~\ref{fig:ar-io}.

The padding rules of \sarseqdata{} are much the same as auto-regressive \seqdata{}. As mentioned in the main submission, we pad the 2-dimensional \sarseqdata{} to $[[y_{1,1}, y_{1,2}, \cdots, y_{1, L}], \cdots,  [y_{M,1}, y_{M,2}, \cdots, y_{M, L}]]$, where $L$ is the maximum length of each sub-sequence and $M$ is the number of sub-sequences. The valid sub-sequences begin with a key-point. For each valid sub-sequence, we follow the same padding rules of \seqdata{}, except there isn't a \texttt{start} token in an input sub-sequence because of the Key-point Prompt. We set the other sub-sequences to the "n/a" class making the loss of these sub-sequences without a key-point not calculated.

\input{figure/AR-io}

Thresholds for Reachability Precision-Recall are chosen from $[0.5, 1.0, 1.5, 2.0, 2.5]m$.

\subsection{Auto-Regressive \name{}}
Based on auto-regressive dependency of \seqdata{}, we design our baseline as auto-regressive ~\seqdata{} generation.

\noindent \textbf{Architecture.} 
We apply the same encoder-decoder architecture as \cite{chen2021pix2seq}.
The encoder is responsible for extracting BEV feature $\mathcal{F}$ from multiple onboard cameras such as Lift-Splat-Shoot~\cite{philion2020lift}.
For decoder, we use the same Transformer decoder as \cite{chen2021pix2seq} which includes a self-attention layer, a cross-attention layer and a MLP layer.

\noindent \textbf{Objective.}
We denote the ground-truth \seqdata{} with length $L$ as $y$ and the predicted \seqdata{} as $\hat{y}$, then the objective of auto-regressive \name{} is maximum likelihood loss, \ie
\begin{equation}
    \max \sum_{i=1}^{L} w_i \log{P(\hat{y}_i|y_{<i}, \mathcal{F})},
\end{equation}
where $y_i$ is the $i^{th}$ token of $y$, $y_{<i}$ means all tokens before $y_i$ and $w_i$ is the class weight.
In practice, since the label \texttt{Lineal} for $v_c$ and index $0$ for $v_d$ appear most frequently, we set $w_j$ as a small value for these class.

\noindent \textbf{Input and target sequence construction.}
The input sequence starts with a \texttt{start} token and the target sequence ends with an \texttt{EOS} token.
We also apply synthetic noise objects technique~\cite{chen2021pix2seq} to the sequence construction.
Details are shown in the Supplementary material. 

\noindent \textbf{Efficiency.}
Suppose the Transformer spends $\mathcal{T}_s$ time inferring a single query.
The inference time complexity should be $\mathcal{O}(|\mathcal{E}|\cdot \mathcal{T}_s)$.

\subsection{Semi-Autoregressive \name{}}
The vanilla Auto-Regressive \name{} generates \seqdata{} one by one, which is a highly time-consuming process. 
The reason for this expensive one-by-one generation is the ingrained auto-regressive dependency assumption.

In the field of Natural Language Processing, the human language is highly cohesive, which means that any attempt to generate text without auto-regressive assumption can result in a significant decrease in accuracy~\cite{xiao2022survey}.
However, it's not the case for road network.
With regards to the Figure~\ref{fig:intro}, observations have been made regarding the dependency of \seqdata{}:
(i) The locations of certain road points (start points, fork points or merge points) can be {\bf independent} of previous vertices and instead depend solely on the BEV feature map, \ie,
\begin{equation}
    P(y_i|y_{<i}, \mathcal{F}) = P(y_i|\mathcal{F}).
\end{equation}
(ii) Except for locations of these road points, other tokens are still strongly auto-regressive.

Drawing from these findings, we suggest the adoption of  Semi-autoregressive \name{} ({\bf\em\sarabb{}}) that retains auto-regressive functionality within local contexts while simultaneously conducting none-autoregressive generations in parallel.
To facilitate this approach, we propose a novel representation of the road graph called {\bf\em \sarseqdata{}}. 

\input{figure/SAR-RTSeq}
\noindent \textbf{Semi-Autoregressive \seqdata{}.}
The objective of Semi-Autoregressive \seqdata{} is to divide the trees in the Directed Forest into smaller sub-trees as much as possible so that each tree will be simultaneously inferred and the auto-regressive length can be reduced as much as possible.
As shown in Figure~\ref{fig:sar-rtseq}, we begin by identifying all key-points in the Directed Acyclic Graph (DAG) that meet the condition $od(v)>1$ or $id(v)>1$ or $id(v)=0$.
We then proceed to recursively extract these points from their original parent and sub-tree until they become roots with an $id(v)$ value of 0.
To restore the edge between the key-point $v$ and its parent, we create a duplicate of the parent and assign it as the child of $v$ with a special label, \texttt{Clone}, identical to that used in \seqdata{}.
Similarly, the \texttt{Clone} will only be traversed after its original vertex is traversed.
As shown in Figure~\ref{fig:sar-rtseq}, different from the auto-regressive \seqdata{}, we construct an independent sequence for each independent tree, so that the SAR-\seqdata{} is a 2-dimension sequence, \ie, $[[y_{1,1}, y_{1,2}, \cdots, y_{1, L}], \cdots,  [y_{M,1}, y_{M,2}, \cdots, y_{M, L}]]$, where $L$ is the maximum length of each sub-sequence and $M$ is the number of sub-sequences.
The padding rules is shown in the Supplementary material.
Noted that the new data structure is also a directed forest, therefore the construction and recovering follow all rules in the auto-regressive \seqdata{}.

\noindent \textbf{Architecture.}
The \sarabb{} can be divided into three parts: (i) Ego-car Feature Encoder, (ii) Key-point Transformer Decoder, (iii) Parallel-Seq Transformer Decoder.
Ego-car Feature Encoder follows that in \arabb{}.
{\em Key-point Transformer Decoder} is a parallel Transformer decoder~\cite{carion2020end}, which takes a fixed set of learned positional embeddings as input and predict locations of key points based on set prediction~\cite{carion2020end}.

Then, {\em Parallel-Seq Transformer Decoder} is proposed for solving mixture of auto-regressive and non-autoregressive problem, \ie.
\begin{equation}
\label{equ:sa}
    \max \sum_{i=1}^M \sum_{j=1}^L P(y_{i, j}\ |\ y_{1:M, 1:j-1}, \mathcal{F}, \mathcal{V}_{kp}),
\end{equation}
where $\mathcal{V}_{kp}$ represents location of key-points detected from Key-points Transformer Decoder.
For a certain $i$, $y_{i,j}$ is generated auto-regressively, while for a certain $j$, all $y_{1:M, j}$ are generated in parallel.
The dependency is illustrates in Figure~\ref{fig:sar-rttr-arc}.

However, following this objective will cost $\mathcal{O}(M^2\times L^2)$ memory complexity for self-attention~\cite{vaswani2017attention}.
Inspired by \cite{wang2020axial}, where a cross combination of self-attention from different directions leads to a final global attention, we design two self-attention applied on different directions of the 2-dimension \sarseqdata{}.
As shown in Figure~\ref{fig:sar-rttr-arc}, {\em Intra-seq self-attention} first applies self-attention on $y_{1:M, j}$ for each $j$ and {\em Inter-seq self-attention} then applies self-attention on $y_{i, 1:j-1}$ for each $i$.
The memory complexity is reduced to $\mathcal{O}(M^2+L^2) \ll \mathcal{O}(M^2\times L^2)$.

\noindent \textbf{Key-point prompt learning.}
To deduce $y_{i,1:L}$ using $\mathcal{V}_{kp}$ as a basis according to Equation~\ref{equ:sa}, we implement Key-point Prompt Learning. 
Key-point Prompt for a sub sequence $y_{i,1:L}$ contains two parts: (i) locations of all key-points; (ii) location of the start key-point (\texttt{Ancestor}) of $y_{i,1:L}$.
As depicted in Figure~\ref{fig:sar-rttr-arc}, 
this involves organizing the locations of key-points and the start key-point location as a sequence of discrete tokens, and assigning dedicated word embeddings and position embedding for the prompt. 
 The Key-point Prompt is then added to Semi-Autoregressive \seqdata{} facilitates the aggregation of key-point information in the sequence.

\noindent \textbf{Objective.}
The objective contains two part: key-points detection and auto-regressive MLE loss.
Key-points are optimized by Hungarian loss~\cite{carion2020end}.
We denote the set of $M$ predictions as $\hat{z}=\{\hat{z}^{(i)}\}_{i=1}^M$, and the ground-truth $z$.
Each $z_i$ composes $z^{(i)}=(c^{(i)}, k^{(i)})$, where $c^{(i)}\in\{0, 1\}$ denotes whether the prediction is a key-point and $k^{(i)}=(k^{(i)}_x, k^{(i)}_y)$ is the position of key-points.
We then build the pair-wise matching cost $\mathcal{L}_{\text{match}}(z_i, \hat{z}_{\sigma(i)})$ between ground-truth $z_i$ and prediction $z_{\sigma(i)}$.
We define the matching cost as class probability and key-points L-1 distance, \ie $\mathcal{L}_{\text{match}}(z_i, \hat{z}_{\sigma(i)})$ as $-\mathbb{1}_{\{c_i\neq 0\}}\hat{p}_{\sigma(i)}(c_i)+\mathbb{1}_{\{c_i\neq 0\}}\|k^{(i)}-\hat{k}^{(i)}\|_1$.
Based on this matching cost, we find a bipartite matching between these two sets with the lowest matching cost.

The second step is to compute the Hungarian loss for all pairs matched in the previous step. 
The Hungarian loss is a linear combination of a negative log-likelihood for class prediction and a L-1 loss.
\begin{align}
    \notag \mathcal{L}_{\text{Hungarian}}(z, \hat{z}) &= \sum_{i=1}^M\left[ -\log{\hat{p}_{\sigma(i)}(c_i)}\right.\\
    &+ \left. \mathbb{1}_{\{c_i\neq 0\}}\|k^{(i)}-\hat{k}^{(i)}\|_1 \right].
\end{align}

Auto-regressive MLE loss follows that of \arabb{}.

\noindent \textbf{Efficiency and dependency.}
Suppose the Transformer spends $\mathcal{T}_s$ time inferring a single query, and the parallel acceleration rate for GPU is $\alpha\ll 1$.
The inference time complexity of combination of Key-point and Parallel-seq Transformer can be approximated as $\mathcal{O}(\alpha(|\mathcal{E}|+|\mathcal{V}_{kp}|)\cdot \mathcal{T}_s)$ which greatly less than that of Auto-Regressive by ratio $\alpha(1+\frac{|\mathcal{V}_{kp}|}{|\mathcal{E}|})$.
In addition to acceleration, improved dependency modeling can also enhance contextual reasoning~\cite{zhu2020deformable, wang2022anchor, bai2022transfusion}, thus benefiting performance.

\input{figure/NAR-RTTR-arc}
\subsection{Non-Autoregressive \name{}}
While the Semi-autoregressive \name{} improves inference speed to some extent, there is still an inefficient auto-regressive component present. 
Therefore, we propose a fully non-autoregressive generation model, called Non-Autoregressive \name{}, which can output the entire sequence at once.
To mimic the auto-regressive generation, we iteratively refine the output from this non-autoregressive generation , \ie, 
\begin{equation}
    \max \sum_{i=1}^M \sum_{j=1}^L P(y_{i, j}\ |\ \hat{y}, \mathcal{F}, \mathcal{V}_{kp}),
\end{equation}
where $\hat{y}$ is the predicted Semi-Autoregressive \seqdata{} of the last guess.
Provided the limited iteration times, the Non-Autoregressive \name{} can achieve a extremly high inference speed than its auto-regressive counterpart.

The training and inference approach are visualized in Figure~\ref{fig:nar-rttr-arc}.
During training, we utilize a masked language modeling strategy~\cite{devlin2018bert} that involves masking a high percentage of the input ground-truth sequence and prompting the Transformer to predict all the missing tokens. 
Thus, during inference, we begin with a fully masked sequence and predict the Semi-Autoregressive \seqdata{} multiple times, with each iteration masking tokens with low confidence.
With each iteration, the results will be gradually refined. 

\noindent \textbf{Efficiency.}
Time complexity for \narabb{} can be approximated as $\mathcal{O}(\alpha |\mathcal{V}_{kp}|(N_{iter}+1)\cdot \mathcal{T}_s)$.
Acceleration ratio is $\frac{|\mathcal{E}| / |\mathcal{V}_{kp}|+1}{N_{iter} + 1}$ where $N_{iter}\ll |\mathcal{E}| / |\mathcal{V}_{kp}|$.

\subsection{Details of RoadNet Sequence Representation}

Our method first formalizes road networks as directed acyclic graphs (DAGs) encoding both Euclidean geometry and non-Euclidean topology (Section III (A)).
These graphs are then serialized into two types of representations:
(i) a coupled format, which binds geometry and connectivity tightly (Section III (B)), and
(ii) a decoupled format, which separates node and edge information for modeling flexibility (Section III (C)).
To decode these sequences, we design three Transformer-based strategies:
(i) an autoregressive decoder (AR-RNTR) that generates tokens sequentially (Section III (E)),
(ii) a semi-autoregressive decoder (SAR-RNTR) that enables partial parallel decoding through keypoint grouping (Section III (F)), and
(iii) a non-autoregressive decoder (NAR-RNTR) that predicts all tokens in parallel and refines them iteratively (Section III (G)).
For the decoding process, we represent each vertex–edge pair using six integers: the vertex location ($v_x$, $v_y$), vertex category $v_c$, index of the parent vertex $v_d$, and the Bezier curve coefficients ($e_{px}$, $e_{py}$).
The details are provided in Table~\ref{tab:embedding}.
Together, as illustrated in Figure~\ref{fig:pip_re_v2}, these components collectively contribute to a cohesive and robust representation that significantly enhances road network construction and decoding process in terms of geometry, topology, and computational efficiency.

Recognizing that perception errors from BEV inputs can propagate and mislead topology generation, we introduce two enhancement modules. The first is a topology-inherited training (TIT) strategy that decouples perception and reasoning by pretraining topology on clean BEV maps, distilling it to noisy inputs, and fine-tuning jointly, as shown in Section III (K). The second enhancement leverages SD-Map sequences as prompts, injecting prior knowledge of road connectivity to guide generation and boost generalization, as shown in Section III (L). 
Together, as in Figure~\ref{fig:pip_re}, these components form a cohesive and robust system that significantly improves road network reconstruction across geometry, topology, and efficiency dimensions.

\input{figure/pipeline_re_v2}
\input{figure/pipeline_re}

\input{figure/metrcis}
\subsection{Metrics}
\noindent \textbf{Precision-recall.}
Precision is defined as 
\begin{equation}
    \text{Precision} = \frac{\text{True Positive}}{\text{True Positive}+\text{False Positive}}.
\end{equation}
Recall is defined as 
\begin{equation}
    \text{Recall} = \frac{\text{True Positive}}{\text{True Positive}+\text{False Negative}}.
\end{equation}
And F1 score is defined as 
\begin{equation}
    \text{F1 score} = \frac{2\times \text{Precision} \times \text{Recall}}{\text{Precision} + \text{Recall}}.
\end{equation}
To ensure a fair comparison, we employ three metrics from \cite{can2021structured, can2022topology}: {\bf\em mean precision-recall}, {\bf\em detection ratio}, and {\bf\em connectivity}.
However, these metrics, which rely solely on centerline detection, neglect the significance of both the location accuracy of road-points and the reachability of the road graph.
To make up with the deflect, we propose 2 following metrics.

\noindent \textbf{Landmark precision-recall.}
We use  landmark precision-recall to evaluate the location accuracy of landmarks. 
For each predicted landmark, we match it to a ground-truth with the nearest distance. 
If a predicted landmark is within the threshold distance with its matched ground-truth, it is true positive, otherwise it is false positive.
If a ground truth landmark is not matched with any predictions or not within the threshold distance with its matched prediction, it is false negative.
Thresholds for landmark precision-recall are chosen from $[0.5, 1.0, 1.5, 2.0, 2.5, $ $3.0, 3.5, 4.0, 4.5, 5.0]m$.

\noindent \textbf{Reachability precision-recall.}
One of the motivations for predicting road topology is finding the valid paths of any two points on the road, which allows the self-driving car to reach its destination. 
Reachability between landmarks is defined that there exists a path going from one landmark to another.
After matching landmarks in the former session, we propose reachability precision-recall to evaluate both connectivity and accuracy of path between landmarks. 
In the road network, if two landmarks are connected, it means there exists a path from one landmark to another. 
Given a pair of predicted landmarks $\hat{A},\hat{B}$ and its matched pair of ground-truth landmarks $A,B$,
a true positive is a path connecting $\hat{A},\hat{B}$ with Chamfer Distance to any of the ground-truth path connecting $A,B$ less than the threshold.
A false negative is a ground truth path from $A,B$ with no matched predicted path. 
Thresholds for reachability precision-recall are chosen from $[0.5, 1.0, 1.5, 2.0, 2.5]m$.
An example is given in Figure~\ref{fig:metrcis}.

%% file: figure/AR-RTSeq.tex
\begin{figure}[tb]
    \centering
    \includegraphics[width=\linewidth]{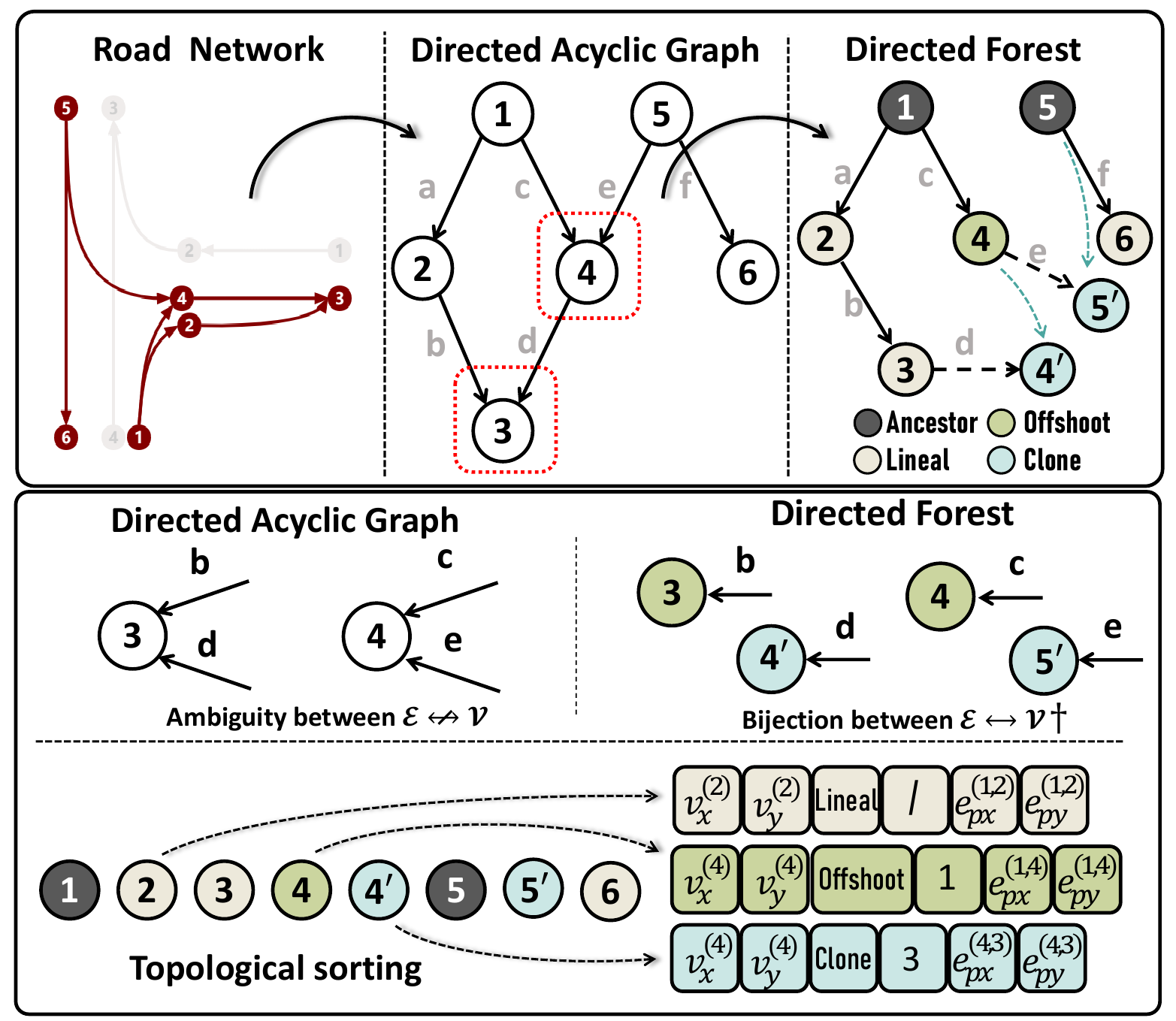}
    \caption{
    {\em Top} illustrates the transformation process from a Directed Acyclic Graph to a Directed Forest for \seqdata{}. 
    The \textcolor{red}{red} boxes enclose all merge-points that have non-unique parents.
    We replicate theirs parents (except the first parent) to their children (\textcolor{cyan}{cyan nodes}) and delete the corresponding edges. 
    {\em Middle} demonstrates a bijection between the edges and vertices ($\dag$: excluding the \texttt{Ancestors}). 
    {\em Bottom} lists the topological sorting result of vertices and presents six integers for each vertex.
    }
    \label{fig:ar-rtseq}
    \vspace{-4mm}
\end{figure}

%% file: figure/SAR-RTTR-arc.tex
\begin{figure*}[tb]
    \centering
    \includegraphics[width=\linewidth]{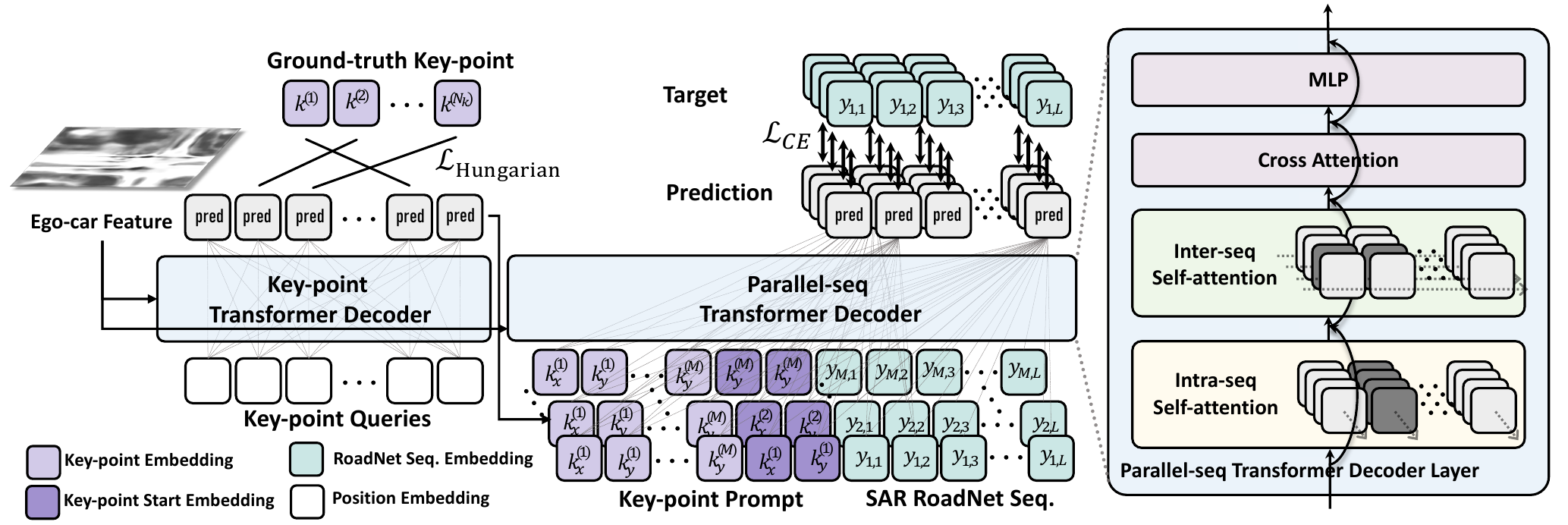}
    \caption{
    {\em Left} shows two major components of Semi-Autoregressive \name{}: Key-point Transformer decoder \revision{to detect key-points} and Parallel-seq Transformer Decoder \revision{to generate Semi-Autoregressive RoadNet Sequence}. ``SAR RoadNet Seq" stands for Semi-Autoregressive RoadNet Sequence.
    {\em Right} decouples self-attention of Parallel-seq Transformer Decoder layer into Intra-seq and inter-seq self-attention.
    }
    \label{fig:sar-rttr-arc}
    \vspace{-4mm}
\end{figure*}

%% file: figure/DAR-RTSeq.tex
\begin{figure}[tb]
    \centering
    \includegraphics[width=\linewidth]{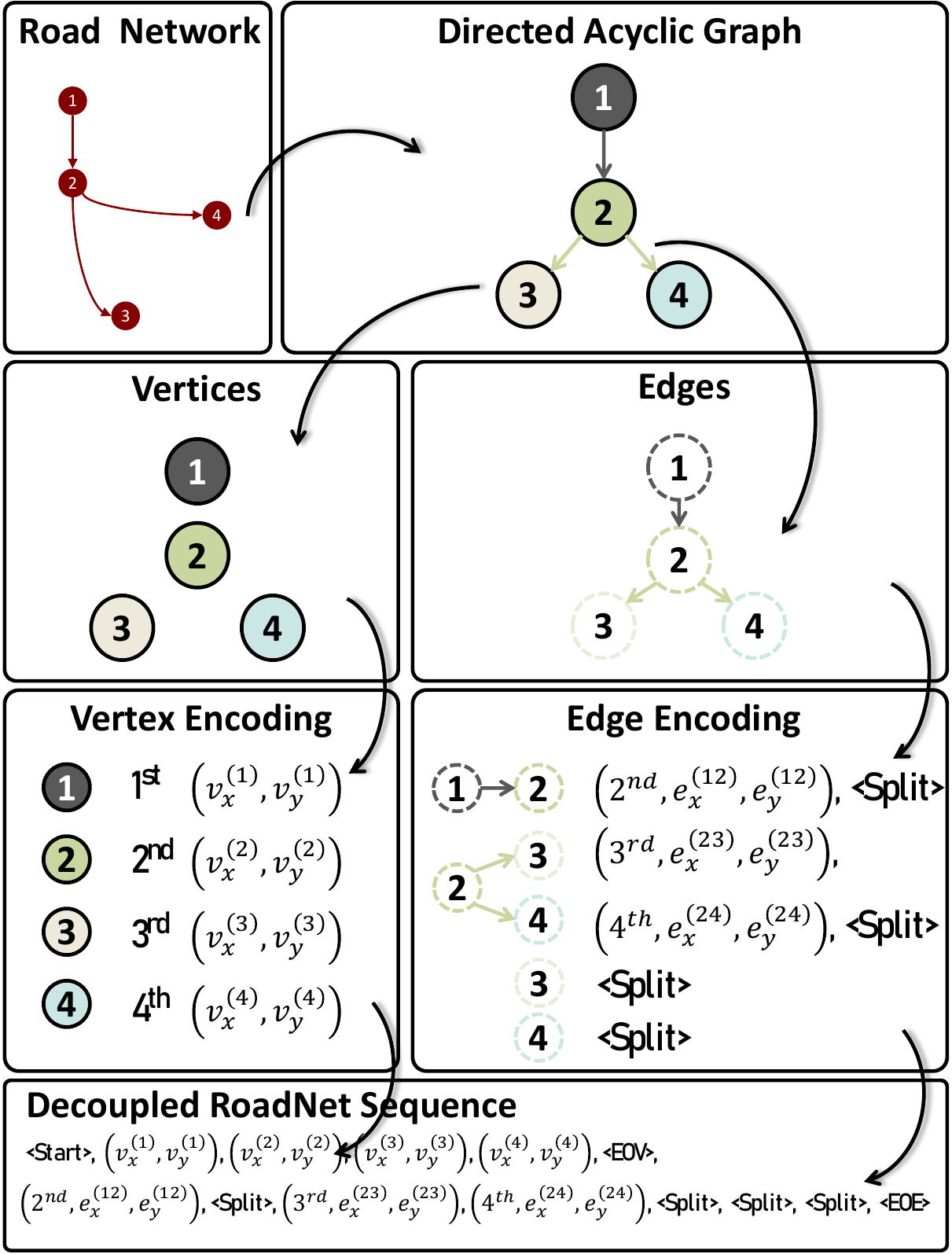}
    \caption{
    \ext{
     The sequence of a decoupled \seqdata{} comprises two components: the \textit{\textbf{vertex sequence}} and the \textit{\textbf{edge sequence}}. 
     The vertex sequence consists of vertex coordinates arranged in depth-first search order and is encoded by their coordinates. 
     For each vertex, the edges starting from it are listed in edge encoding in order. Finally, the vertex sequence and edge sequence are concatenated with special tokens to indicate the start and end.}
    }
    \label{fig:dec_ar-rtseq}
    \vspace{-6mm}
\end{figure}

%% file: table/embedding.tex
\begin{table}[htb]
  \centering
  \caption{Embedding range of different integers.}
    \begin{tabular}{ lc}
    \hline
    
    \hline
    Item & Range\\
    \hline
    $v_x, v_y$ & $0\sim199$\\
    $v_c$ & $200\sim249$\\
    $v_d$ & $250\sim349$\\
    $e_{px}, e_{py}$ & $350\sim569$\\
    \texttt{noise category} & $570$\\
    \texttt{EOS} & $571$\\
    \texttt{Start} & $57$2\\
     \texttt{n/a} & $573$\\
     
    \hline

    \hline
    \end{tabular}
    
  \label{tab:embedding}
\end{table}

%% file: figure/AR-io.tex
\begin{figure}[tb]
    \centering
    \includegraphics[width=\linewidth]{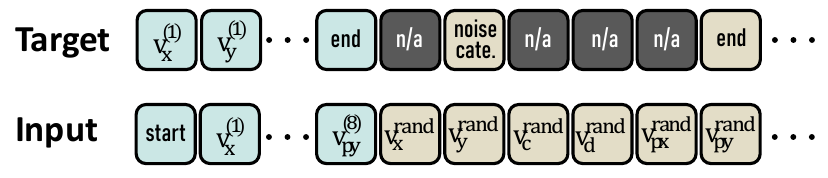}
    \caption{
    An illustration of synthetic noise objects technique~\cite{chen2021pix2seq} on \seqdata{}.
    Loss weight for \texttt{n/a} tokens are set to zero.
    \texttt{Noise cate.} stands for noise category.}
    \label{fig:ar-io}
    \vspace{-6mm}
\end{figure}

%% file: figure/SAR-RTSeq.tex
\begin{figure}[tb]
    \centering
    \includegraphics[width=\linewidth]{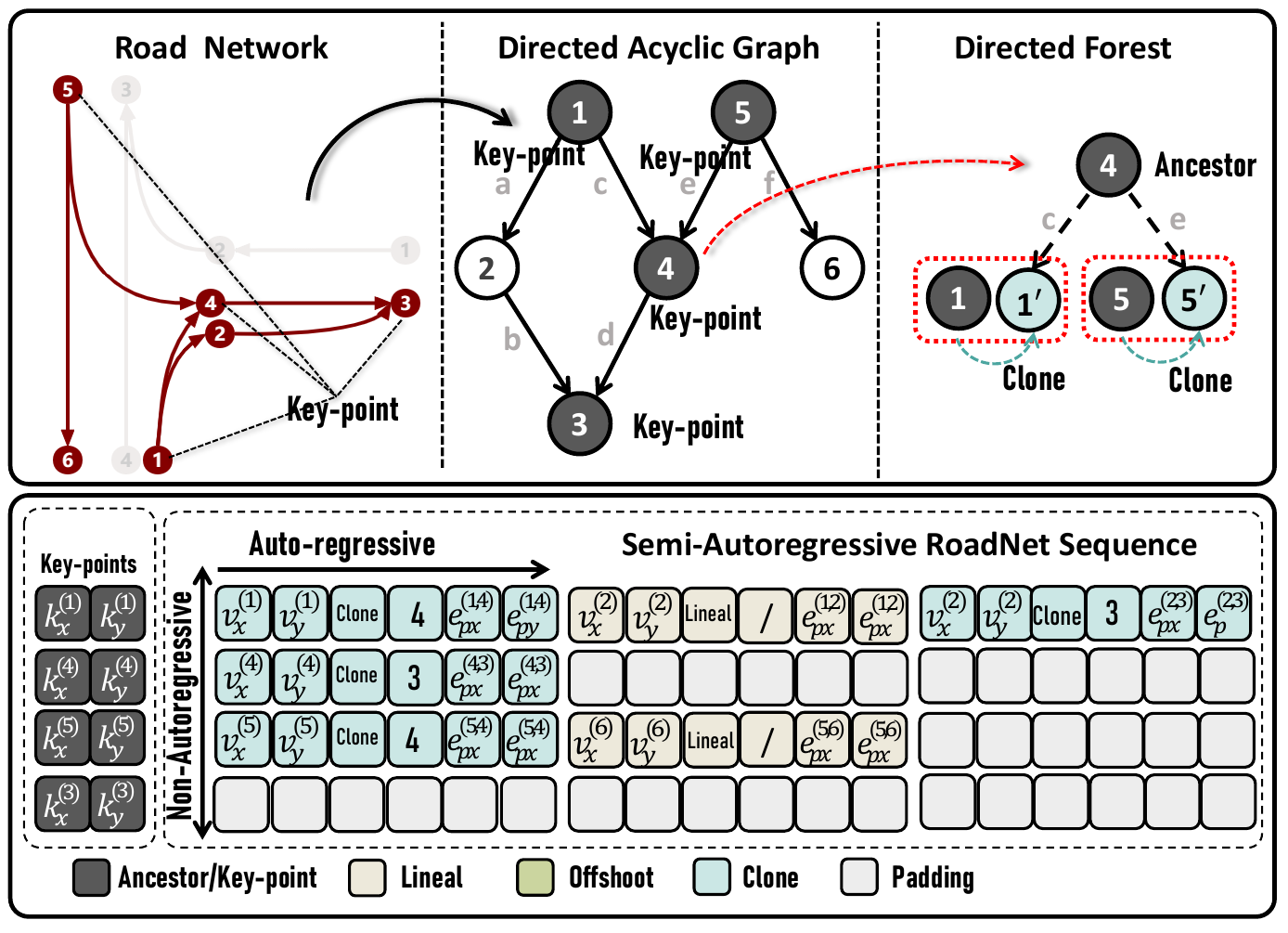}
    \caption{
    \revision{
    {\em Top} indicates the selection of key-points and the approach to construct sub-trees taking these key-points as root.}
    {\em Bottom} presents the \sarseqdata{} for the example mentioned above.
    }
    \label{fig:sar-rtseq}
    \vspace{-6mm}
\end{figure}

%% file: figure/NAR-RTTR-arc.tex
\begin{figure}[tb]
    \centering
    \includegraphics[width=\linewidth]{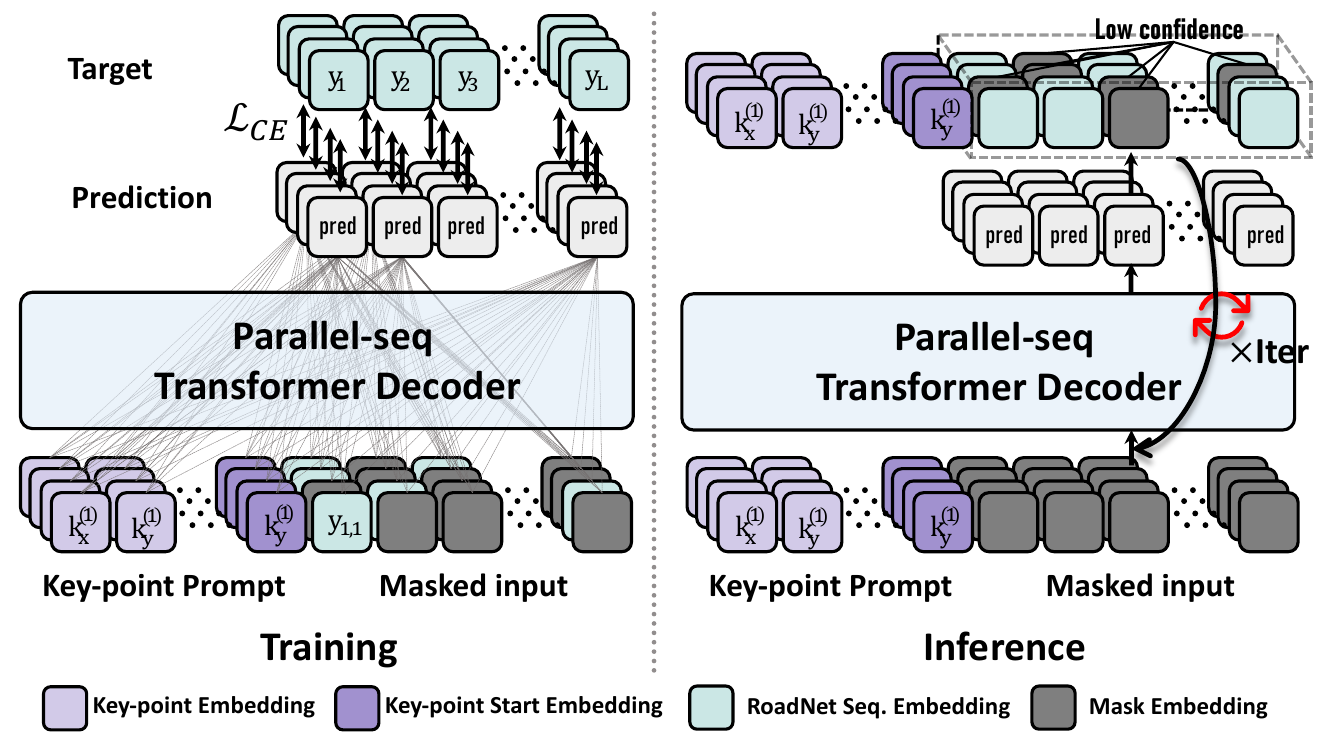}
    \caption{
    {\em Left} illustrates the training stage which takes masked (gray square) \sarseqdata{} as input.
    {\em Right} shows inference stage where the Parallel-seq Transformer Decoder takes a fully-masked sequence as input and iteratively masks the predicted token with low confidence.
    }
    \label{fig:nar-rttr-arc}
    \vspace{-6mm}
\end{figure}

%% file: figure/pipeline_re_v2.tex
\begin{figure}[tb]
    \centering
    \includegraphics[width=0.85\linewidth]{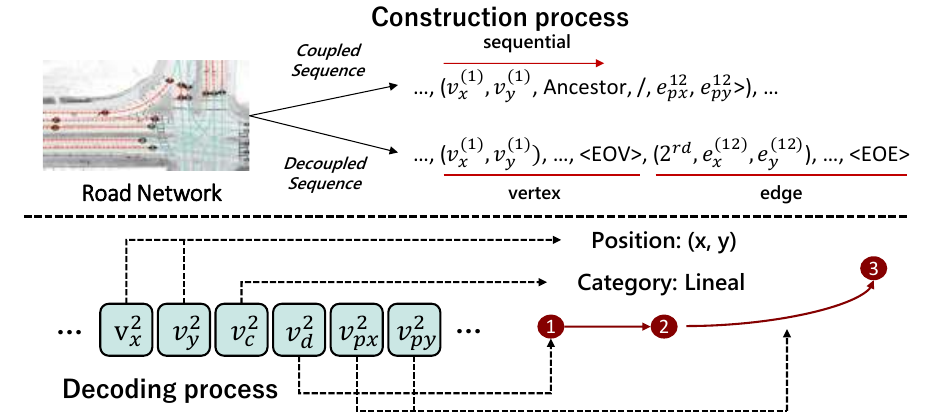}
    \caption{
    A visual explanation of RoadNet sequence construction and decoding process.}
    \label{fig:pip_re_v2}
\end{figure}

%% file: figure/pipeline_re.tex
\begin{figure}[tb]
    \centering
    \includegraphics[width=0.85\linewidth]{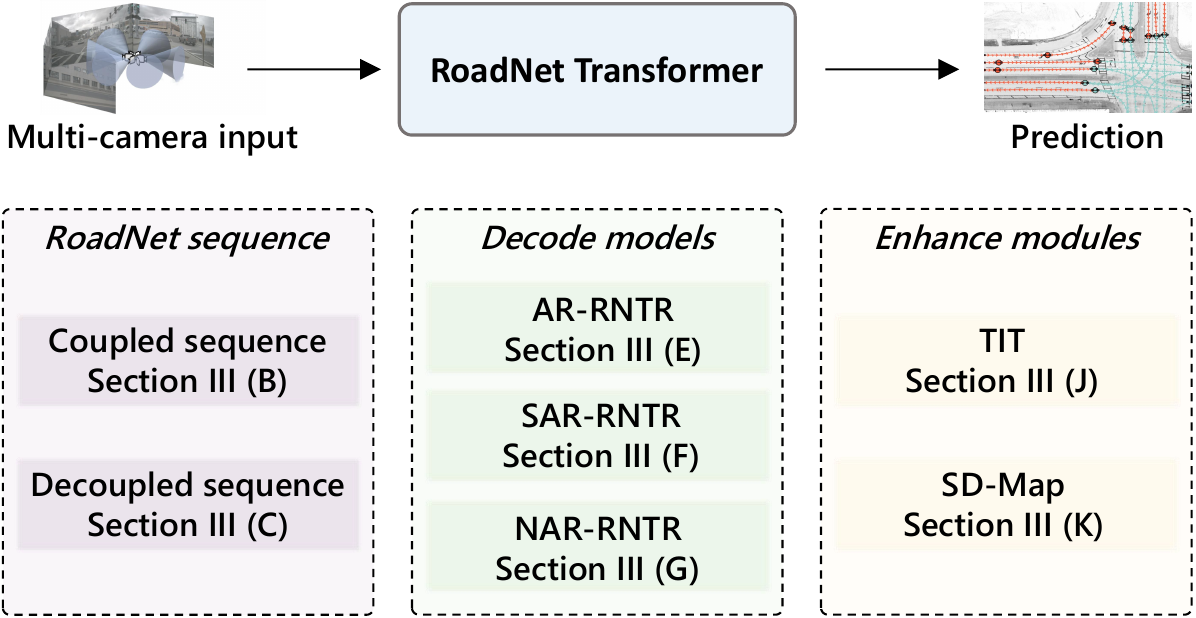}
    \caption{
    A visual illustration of the entire method, including the proposed RoadNet sequence, the three models, and the enhancement modules used to improve performance.}
    \label{fig:pip_re}
    \vspace{-6mm}
\end{figure}

%% file: figure/metrcis.tex
\begin{figure}[tb]
    \centering
    \includegraphics[width=\linewidth]{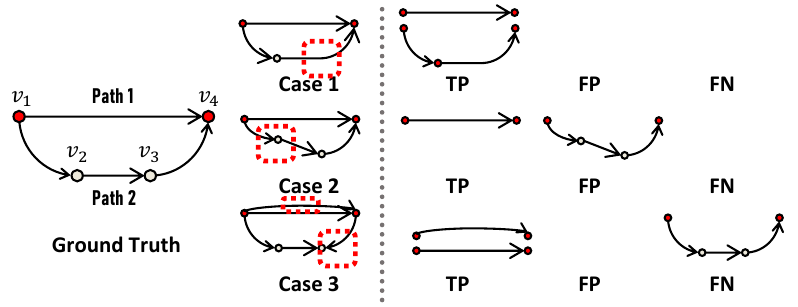}
    \caption{Examples of TP, FP and FN when evaluating reachability between $v_1$ and $v_4$. 
    Although case 1 misses landmark $v_2$, but both paths between $v_1$ and $v_4$ are within distance of the ground-truth paths.
    Case 2, however, has path 2 surpasses the threshold distance from matched its matched ground-truth path 2 so it's false positive.
    Case 3 predicts path 1 twice but they are both within threshold distance.
    However, path 2 in the ground-truth has no matched prediction so it's false negative.
    }
    \label{fig:metrcis}
\end{figure}

%% file: file/3-new-method.tex
\input{figure/bottleneck}
\input{figure/pretrain}
\subsection{Bottleneck of Sequence-to-sequence Methods}

\ext{After proposing our auto-regressive, semi-autoregressive, and non-autoregressive sequence-to-sequence methods for road network extraction, we examine the bottlenecks of the sequence-to-sequence approach. 
First, since the default split of the nuScenes dataset includes overlapping scenes in the training and evaluation sets, we need to create a new split to ensure non-overfitting. 
Unlike before~\cite{roddick2020predicting}, we should also ensure that different daytime and weather conditions are equally distributed in the training and evaluation sets.
We then retrain the road segmentation model on the new split dataset and train the \arabb{}. The visualization of the results is presented in Figure~\ref{fig:bottleneck}.
Clearly, occlusion in the round-view camera cannot be avoided, making accurate road perception impossible even in ideal conditions. For example, the far side of a crossroad or a U-turn cannot be predicted as they are invisible. Therefore, in the \arabb{} results, the inaccurate Euclidean understanding impairs the perception of non-Euclidean data, meaning location errors propagate to the topology learning.
}

\ext{
Since the Euclidean understanding is primarily the responsibility of the BEV encoder, decoupling the training of the BEV encoder and \name{} is a natural choice to mitigate error propagation during training. 
First, \name{} is trained to enhance its topological understanding, and then we transfer the topology knowledge to both the BEV encoder and \name{}. 
We name this new training strategy {\bf\em \strategy{}} ({\bf\em \stabb{}}).
}

\subsection{\strategy{}}
\ext{
As mentioned earlier, the Euclidean understanding is primarily the responsibility of the BEV Encoder, so we need to decouple the training of the BEV Encoder and \name{}. 
However, the training of \name{} also depends on the BEV features as input. 
Therefore, we utilize the Lidar map provided by the nuScenes~\cite{caesar2020nuscenes} dataset, which contains high-resolution Birds-Eye View (BEV) road images as an intermediate step.
As shown in the first column of Figure~\ref{fig:tit_arc}, we first use a simple ResNet~\cite{he2016deep} as the BEV encoder for the BEV Lidar map, and \name{} uses the clear BEV features as input to train the \seqdata{}.
}

\ext{
Since the first stage is crucial and much more efficient than using round-view images as input, we utilize the entire nuScenes~\cite{caesar2020nuscenes} dataset with a 20Hz sample rate instead of the 2Hz rate. This approach increases the scale of data to a factor of nine.
}

TIT decomposes the original objective function
\begin{equation}
    \min_{\theta_f, \theta_g} \mathbb{E}_{(x, y) \sim D} \mathcal{L}_{\text{CE}}(g(f_I(x)), y),
\end{equation}
into a two-step optimization process:
\begin{equation}
    \min_{\theta_g} \mathbb{E}_{(x_L,y) \sim D_L} \mathcal{L}_{\text{CE}}(g(f_L(x_L)), y),
\end{equation}
\begin{equation}
    \min_{\theta_f} \mathbb{E}_{(x, x_L) \sim D} \| f_I(x) - f_L(x_L) \|_1.
\end{equation}
This progressive optimization strategy mitigates gradient oscillations caused by directly optimizing a complex objective, thereby enhancing stability and accelerating overall convergence.

\ext{
In the second stage, as shown in the second column of Figure~\ref{fig:tit_arc}, we train the BEV Encoder with round-view image input using distillation. 
We freeze the parameters of the BEV Encoder when using the Lidar map input and keep the parameters open for training when using the round-view images input. We employ L-1 loss, so the loss function $\mathcal{L}_{dist}$ can be expressed as:
}

\begin{equation}
    \mathcal{L}_{dist} = \|\mathcal{F}_{img} - \mathcal{F}_{lidar}\|_1.
\end{equation}

Compared to directly optimizing the BEV encoder using cross-entropy loss for the final objective, L1 loss provides more stable gradients, mitigating gradient oscillations caused by topology prediction errors in end-to-end training.
This gradient smoothness ensures more stable parameter updates for the BEV encoder, reducing training instability caused by large gradient fluctuations.

\ext{
In the final stage, we assemble the BEV Encoder with round-view image input and \name{} as in the traditional \arabb{} method, and train for a few epochs to fine-tune the model.
}

\noindent \textbf{Analysis.}
To minimize error propagation in topology reasoning, we adopt a three-stage approach. In the first stage, we use Lidar data to generate accurate BEV features, which are then used to train the RoadNet Transformer for RoadNet Sequence learning. 
In the second stage, we distill LiDAR-based BEV features to camera-based BEV features to refine the encoder's Euclidean understanding.
Finally, in the third stage, we integrate the camera-based BEV features with the RoadNet Transformer for joint learning.
This decoupling allows each component to focus on its specific task—accurate BEV feature generation and RoadNet Sequence learning—effectively reducing error propagation throughout the training process.

The BEV Encoder faces several challenges that impact its effectiveness in complex driving scenarios. First, due to its fixed-resolution grid representation, it may miss fine details of road networks, particularly in dense urban areas where lane markings and road boundaries are closely packed. 
The discretization process results in information loss, which can degrade spatial precision and reduce the encoder’s ability to capture fine-grained structural and semantic features necessary for accurate perception and planning. 
Second, the encoder struggles with non-uniform road layouts, especially in multi-level structures like overpasses and tunnels. Since BEV projections compress 3D spatial information into a 2D plane, it becomes difficult to distinguish overlapping roads, leading to potential connectivity errors in topology reasoning. Third, the reliance on 2D projections limits the encoder’s ability to interpret depth and elevation, making it less effective on uneven terrains. To mitigate these issues, enhancements such as higher-resolution BEV features, multi-scale feature aggregation, and 3D-aware representations could improve accuracy and generalization.

\input{figure/sd_extract}
\input{figure/sdmap_eg}

\subsection{Standard Definition Map (SD-Map)}
\ext{
Although \stabb{} partly solves the bottlenecks, the obstacles of roads from barriers or terrain shown in Figure~\ref{fig:bottleneck} are not fundamentally resolved. Therefore, to address this problem, some prior information should be provided. The prior information should be easy to obtain and inexpensive to process.}

\ext{
In daily driving experiences, navigation maps like Google Maps~\cite{googlemaps} are almost a necessity worldwide, making it easy to gather some information from them. As shown in Figure~\ref{fig:sd_intro_eg}, information in these navigation maps is abundant and difficult for algorithms to decode. Therefore, we extract only road topology information from the navigation map, commonly referred to as standard definition maps (SD-Maps), as shown in the middle of Figure~\ref{fig:sd_intro_eg}.
}

\ext{
The comparison between SD-Map and \data{} is shown in Figure~\ref{fig:sdmap_eg}. Unlike \data{}, which provides lane-level and centimeter-level annotations, SD-Maps contain only road-level and meter-level information. This means they cannot provide the accurate number of lanes for each road, and the locations of landmarks are not precise. Additionally, the road-level annotations imply that some edges between vertices are bi-directional, which means the \data{} cannot be abstracted as a Directed Acyclic Graph (DAG) since cycles may exist.
}

\noindent \textbf{Collection of SD-Map.}
\ext{
We collect SD-Map data from OpenStreetMap~\cite{openstreetmap}, which is a free and open geographic database based on freely licensed geodata sources. We export Tracktopo data from the database, whose raw data appears as shown in the second step of Figure~\ref{fig:sd_extract}. The raw SD-Map is constructed from nodes and their connections with direction, which is very similar to \data{}.
Next, we align the SD-Map with the nuScenes map. Since the nuScenes map does not provide GPS coordinates, we must align them manually. We also manually adjust some nodes to correct obvious errors in the alignment. It should be noted that this human intervention is due to the deficiencies of the nuScenes dataset. In practice, acquiring SD-Map information from navigation maps is easy and efficient. Since we do not expect SD-Maps to provide highly accurate locations, some location errors are acceptable.
}

\noindent \textbf{Construction of \sdseqdata{}.}
\ext{
Since the SD-Map is also constructed with vertices and edges, we can similarly project SD-Maps to sequences as we have done with \seqdata{}. 
The only difference is that SD-Maps may not be Directed Acyclic Graphs since cycles may exist. Therefore, we first need to transform the cyclic graph into a DAG.
To achieve this, we duplicate the vertex whenever it is traversed again during a Depth-First Search. The new vertex will copy the location of its original vertex and be noted as the leaf vertex of the last traversed vertex. 
After being transformed into a DAG, we can simply use coupled or decoupled \seqdata{} to obtain what we call \sdseqdata{}.
To be noted, to reduce the length of \sdseqdata{}, we do not include the Bezier curve control points and only treat edges as straight lines.
}

\noindent \textbf{\toponame.}
\ext{
Based on the success of Auto-regressive \name{} (\arabb{}), which treats road network extraction as a sequence-to-sequence approach, we can simply upgrade to {\bf\em \toponame{}} ({\bf\em \topoabb{}}). 
We use \sdseqdata{} as a sequence prompt before the start of \seqdata{}. 
This change does not alter the \arabb{} model. 
Without modifying the structure of \arabb{} or undergoing complex data processing for a new kind of data, \name{} can easily utilize the SD-Maps, resulting in a significant performance boost.
}

%% file: figure/bottleneck.tex
\begin{figure}[tb]
    \centering
    \includegraphics[width=\linewidth]{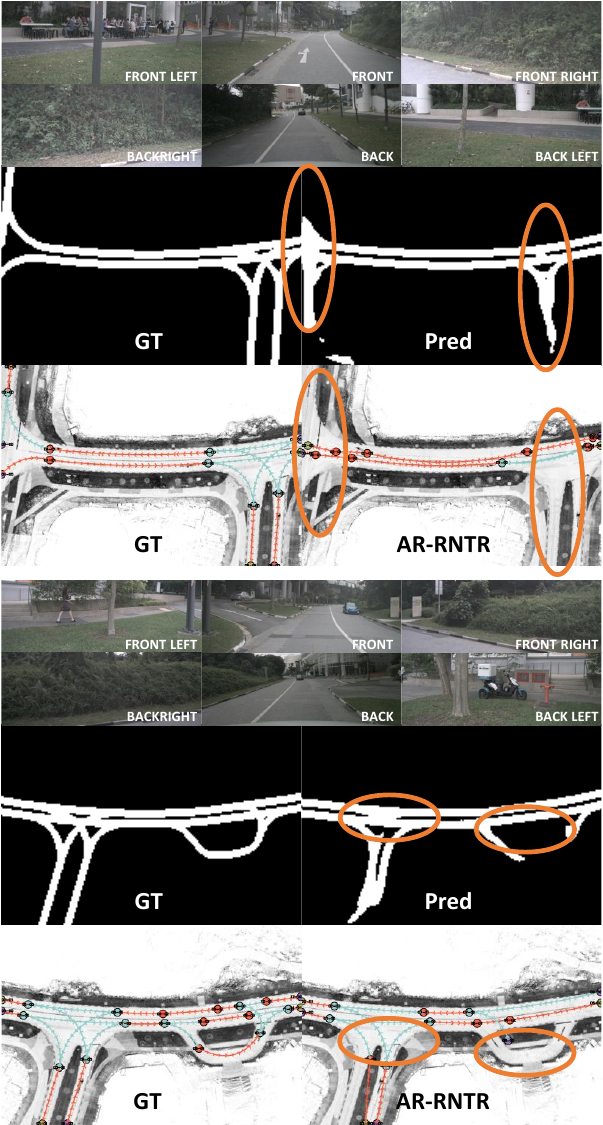}
    \caption{
    \ext{
    Visualization of road segmentation and \data{} extraction results on the new split of the nuScenes~\cite{caesar2020nuscenes} dataset.
{\em Top:} The misprediction of a crossroad obstructed by a barrier impairs the \arabb{}'s ability to predict its topology (circled in orange).
{\em Bottom:} The misprediction of a U-turn and T-road obstructed by terrain also harms the topology prediction.}}
    \label{fig:bottleneck}
    \vspace{-6mm}
\end{figure}

%% file: figure/pretrain.tex
\begin{figure}[tb]
    \centering
    \includegraphics[width=\linewidth]{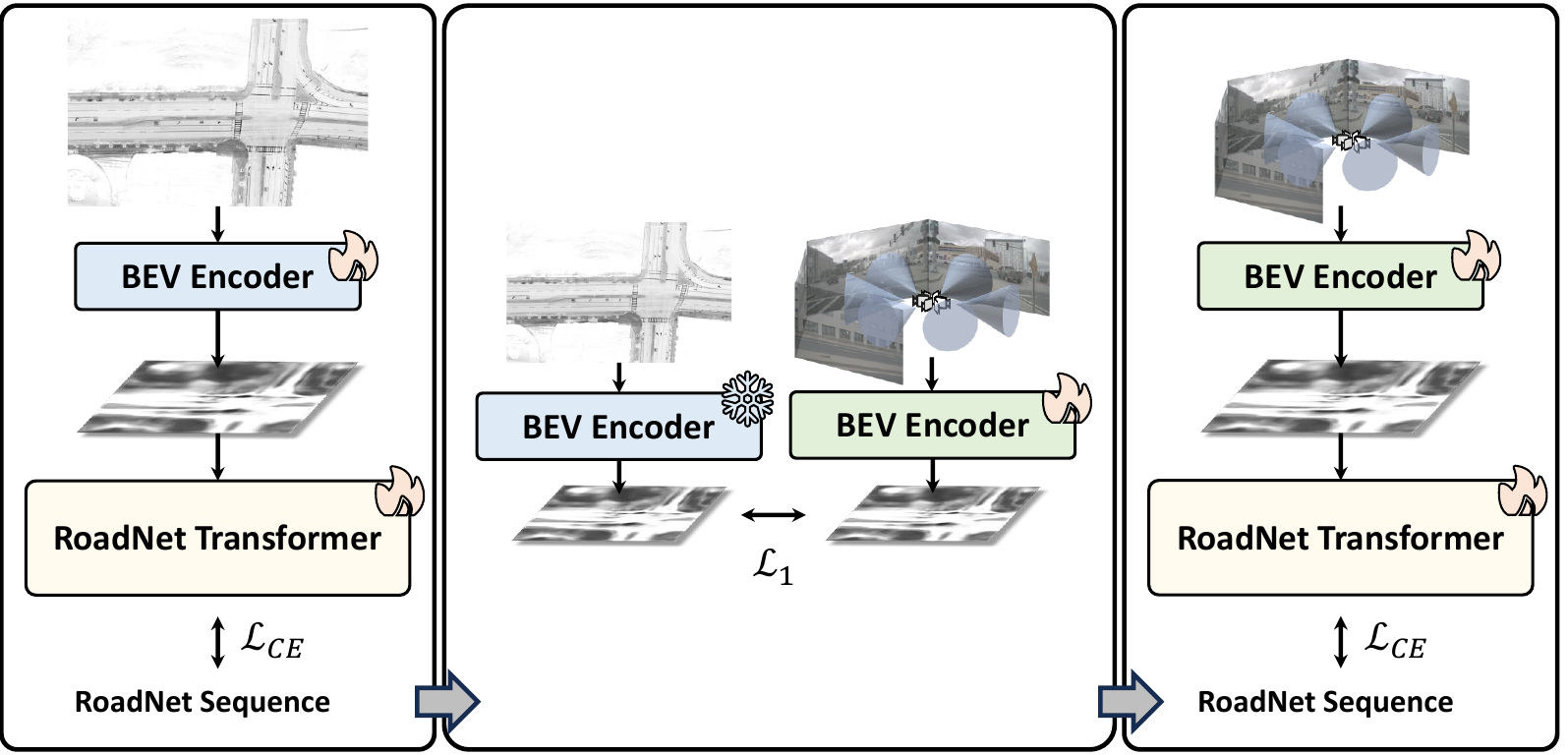}
    \caption{
    \ext{
    {\em Left:} In the first step, a simple ResNet~\cite{he2016deep} is used as the BEV Encoder to encode the BEV Lidar map into BEV features, and \name{} is trained using the clear BEV features from the Lidar map.
    {\em Middle:} In the second step, the BEV Encoder with the Lidar map input is frozen, and we distill the BEV Encoder using round-view images as input to match the previously obtained BEV features.
    {\em Right:} Finally, the BEV Encoder using round-view images as input and \name{} are assembled and fine-tuned for a few epochs.}
    }
    \label{fig:tit_arc}
    \vspace{-2mm}
\end{figure}

%% file: figure/sd_extract.tex
\begin{figure*}[tb]
    \centering
    \includegraphics[width=\linewidth]{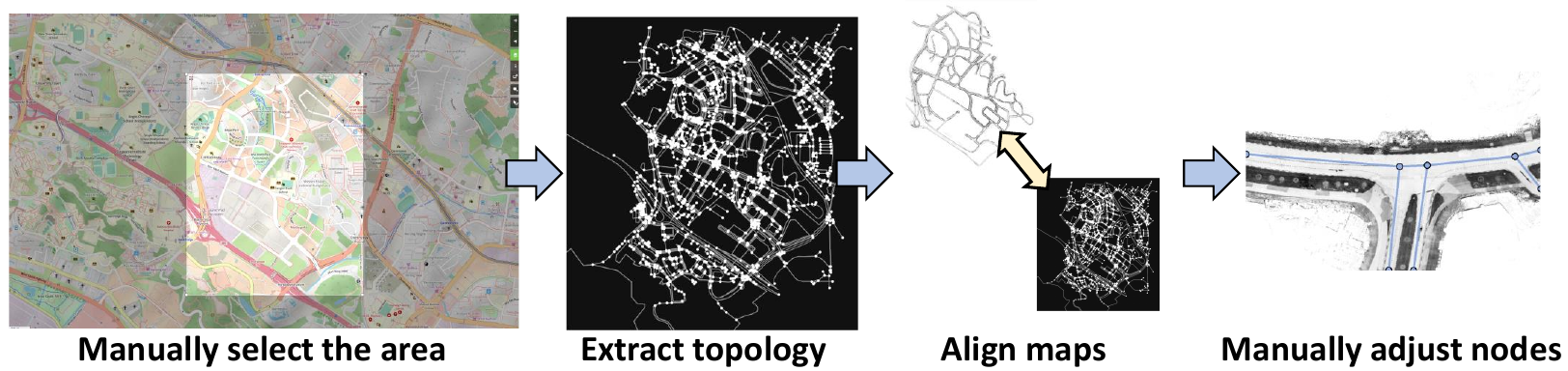}
    \caption{
    \ext{
    Illustration of the SD-Map data collection pipeline. First, we select the area on OpenStreetMap~\cite{openstreetmap} and export its topology data. Then, we manually align the SD-Map with the nuScenes~\cite{caesar2020nuscenes} map. Since the two maps have some local distortions, we manually adjust some nodes to achieve approximate local alignment.}
    }
    \label{fig:sd_extract}
    \vspace{-4mm}
\end{figure*}

%% file: figure/sdmap_eg.tex
\begin{figure}[tb]
    \centering
    \includegraphics[width=\linewidth]{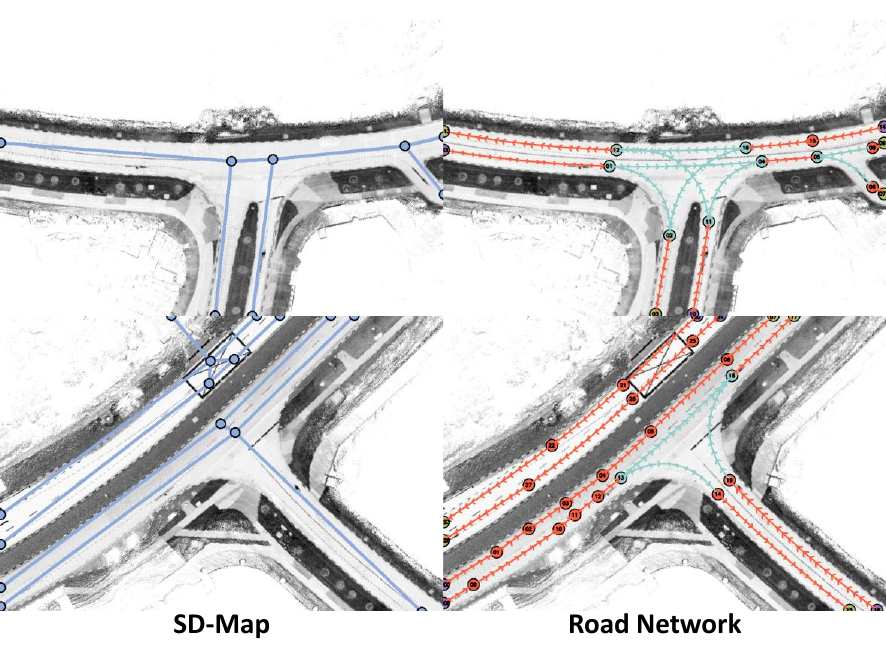}
    \caption{
    \ext{
    Comparison between SD-Map and \data{} in several examples.}
    }
    \label{fig:sdmap_eg}
    \vspace{-4mm}
\end{figure}

%% file: file/4-exp.tex
\section{Experiments}
\subsection{Dataset}
We utilize the nuScenes~\cite{caesar2020nuscenes} dataset (700/150/150 for training/validation/test) to assess our approach. Regardless of the split type, the number of samples in each set remains unchanged. We only utilize sensor information from six cameras, IMU, and GPS, which were sampled at a rate of 2Hz.
\ext{When training \name{} in the first stage of \strategy{}, we utilize sensor information from IMU and GPS sampled at a rate of 20Hz.}

To demonstrate the generalization ability of our approach, we further evaluate it on the Argoverse~\cite{chang2019argoverse} dataset.
The Argoverse dataset is a large-scale autonomous driving benchmark comprising diverse multimodal data collected from real-world urban driving in Pittsburgh and Miami. It contains 113 sequences with over 290,000 annotated 3D object tracks, covering approximately 5 million object instances. The dataset is complemented by high-definition maps spanning nearly 290 kilometers of road networks, which include lane geometry, drivable areas, and traffic semantics, enabling rich contextual modeling.

\subsection{Implementation}

\noindent \textbf{Pretrain.}
We follow LSS~\cite{philion2020lift} for BEV Encoder.
The input images are resized to $128\times352$, and the target BEV area is from -48 to 48m in x-direction (roll) and -32 to 32m in y-direction (pitch) with resolution $1$m in the ego coordinate system. 
ResNet-50~\cite{he2016deep} or VoVNetV2~\cite{lee2020centermask} are used as the image backbone.
The ResNet-50~\cite{he2016deep} model is pretrained using DeepLabV3+~\cite{chen2018encoder} on the Cityscapes dataset\cite{cordts2016cityscapes}, and the VoVNetV2 is pretrained on FCOS3D~\cite{wang2021fcos3d}.
To be noted that the VoVNetV2 is pretrained on the default split of the nuScenes~\cite{caesar2020nuscenes} dataset; therefore, we will not use the VoVNetV2 in the non-overfitting split.
Initially, we pretrain the BEV encoder on centerline segmentation, following all the training strategies outlined in LSS~\cite{philion2020lift}.
During the training of \name{}, we load and freeze the parameters of the BEV encoder. 

\noindent \textbf{\name{}.} Details of three variants are shown below.
{\bf\em\arabb{}: }
The length of \seqdata{} is padded to $6\times 100$.
Transformer decoder layers of \arabb{} are set as 6.
Our \arabb{} is trained on 300 epochs with learning rate $2\times10^{-4}$, batch size as $2\times8$. 
During the training process, we apply random flip, random rotation, and random scaling on BEV feature similar to~\cite{huang2021bevdet,lu2022learning}.
{\bf\em\sarabb{}: }
Transformer decoder layers of \sarabb{} are set as 6 (Key-point) plus 3 (Parallel-seq).
We use noise vertices to pad all sub-sequences to the length of $6\times18$, and the max number of key-points is set to 34. 
The strategies for data augmentation and training are the same as \arabb{}.
{\bf\em\narabb{}: }
During training, we load parameters of trained \sarabb{} and then finetune it with a mask language modeling strategy for 100 extra epochs. 
During inference, the iteration number is set to 3.

\noindent \textbf{\strategy{}.}
\ext{
We apply \strategy{} only to \arabb{}. The basic settings of \arabb{} are the same as those previously introduced for \name{}.
In the first stage, we train on 20Hz data for 300 epochs. We use more data augmentation than when training \name{}. The random rotation is increased from $-45^\circ$ to $45^\circ$, and the scale is adjusted from 0.9 to 1.1.
In the second stage, we train for 24 epochs.
In the third stage, we train for 100 epochs.
}

\noindent \textbf{SD-Map.}
\ext{
We locate \texttt{Singapore onenorth}, \texttt{Singapore Queenstown}, \texttt{Singapore Hollandvillage}, and \texttt{Boston Seaport} from OpenStreetMap~\cite{openstreetmap} and approximately choose the corresponding areas. After exporting the topology data, we retain only the nodes of \texttt{highway} and their secondary type of \texttt{link roads}.
}

\noindent\textbf{Metrics.} Thresholds for Mean Precision-Recall and Landmark Precision-Recall are uniformly sampled from $[0.5, 5.0]$ m.
Thresholds for Reachability Precision-Recall are uniformly sampled from $[0.5, 2.5]$ m.
We only take into account the reachability between vertices that are within a maximum of 5 edges of connection.

\subsection{Results}
\input{table/nus_stsu}
\input{table/ago}
\noindent \textbf{Comparison with state of the art.}
We compare our model with previous state-of-the-art methods on high-definition road network topology extraction.
To achieve a fair comparison, we only utilize {\bf front-view} images as input and ResNet-50~\cite{he2016deep} trained on ImageNet-1K~\cite{russakovsky2015imagenet} as the backbone.
Also, we utilize the PON nuScenes train/val split~\cite{roddick2020predicting} as also applied in \cite{can2021structured, can2022topology}. 
Table~\ref{tab:nus-stsu} presents the results of our approach on the nuScenes dataset using the Mean Precision-Recall, Detection, and Connectivity metrics~\cite{can2021structured}.
On the one hand, our model outperforms all three variants across all metrics, demonstrating the superior performance of \abb{} in both centerline detection and centerline association estimation. 
This remarkable improvement on both centerline location and connectivity can be attributed to the unified representation provided by \seqdata{} and the global context reasoning capabilities of the Transformer architecture.
We also validate our method on Argoverse dataset
in Table~\ref{tab:re_ago}.
The experimental results demonstrate that our method consistently outperforms the competitors.

On the other hand, unlike in Natural Language Processing, our Semi-Autoregressive and Non-Autoregressive versions significantly outperform the Auto-Regressive version, highlighting our dependency decoupling.
\input{table/nus_reach}

\noindent \textbf{Landmark and Reachability.}
Table~\ref{tab:nus-reach} evaluate our methods on Landmark Precision-Recall and Reachability Precision-Recall.
The auto-regressive \abb{} performs worst in most metrics without saying it's slowest inference speed.
In comparison, the Semi-Autoregressive \abb{} leads in all metrics, and boosts the inference speed by 6.0 times.
Remarkably, the non-autoregressive version of \abb{} outperforms the \arabb{} in all metrics and dramatically improves inference speed by a factor of {\bf 47}$\times$, surpassing the nuScenes camera sampling frequency of 2Hz and enabling real-time inference.
The small gap between \narabb{} and \sarabb{} proves the feasibility of the masked sequence training and the iterative based inference.
To summarize, the \sarabb{} achieves the best performance, while the \narabb{} strikes a very balance between efficiency and accuracy.

\input{table/nus_reach_newsplit}

\noindent \textbf{Over-fitting.}
\ext{
Table~\ref{tab:nus-reach-new} shows results with the new split on the nuScenes dataset. Compared to the default split, the results from the non-overlapping split show a significant reduction in both Landmark and Reachability metrics.
The decrease in Landmark performance aligns with the observation that without overfitting, invisible roads cannot be predicted through round-view images. 
Comparing the three variants of \name{}, the Landmark scores show no significant differences, all hovering around 36.5, suggesting that landmark detection has reached its upper limit.
In non-overlapping scenes, the primary challenge is the detection of roads, which leads to less accurate landmark detection compared to the overfitting scenario. This inaccuracy also adversely affects Reachability.
}

\noindent \textbf{\strategy{}} (\textbf{\stabb{}}.)
\ext{
After applying \strategy{}, \arabb{} experiences a significant improvement in both the overfitting and non-overfitting splits. 
In the non-overfitting split, \stabb{} helps \arabb{} exceed the performance of both \sarabb{} and \narabb{}.
Especially in Reachability, \stabb{} leads \sarabb{} by 11.6, demonstrating that with the help of topology data inheritance, topology training is more effective.
In terms of Landmark, however, the improvement is not as obvious, with less than a 3-point increase from \arabb{}. This indicates that \stabb{} cannot fundamentally improve road landmark detection.
}

\noindent \textbf{SD-Maps.}
\ext{
The usage of SD-Maps as prior information brings significant improvements for both the overfitting and non-overfitting splits, with more pronounced benefits for the non-overfitting split. It raises the Landmark F-score from 39.1 to 54.7 and the Reachability F-score from 49.8 to 62.6.
The most notable improvement in the Landmark F-score demonstrates the effectiveness of SD-Maps. Additionally, better landmark detection supports improved topology detection.
}

\subsection{Ablation Studies}
We conduct ablation studies on Transformer layer numbers, as well as mask ratio and iteration times of \narabb{} during inference.
\input{table/ablation_layer}

\noindent \textbf{Number of Transformer layers.}
We investigated the impact of the number of Transformer layers on \narabb{} in Table~\ref{tab:abl_layers}.
Using fewer Key-point Transformer decoder layers leads to a significant loss in accuracy for both landmark and reachability, but provides a limited speedup.
Conversely, using fewer Parallel-seq Transformer decoder layers dramatically improves inference speed while incurring less accuracy loss.

\noindent \textbf{Intra-Seq self-attention.}
The bottom row of Table~\ref{tab:abl_layers} highlights the crucial role of Intra-seq self-attention. Its absence results in a significant loss in both landmark localization and topology connection.

\input{table/ablation_mask_iter}

\noindent \textbf{Mask ratio and iteration times.}
Our investigation into the impact of the mask ratio during the training of \narabb{} revealed the importance of using a large mask ratio. 
Consequently, a 90\% mask ratio is used during Non-Autoregressive finetuning.
We also observed accuracy saturation as the number of iterations increased. 
Therefore, we used 3 iterations for our approach.

\input{table/ablation_aug}
\input{table/ablation_weight}
\input{table/ablation_tit}
\input{table/param_finetune}
\input{table/standard_deviation}
\input{table/nusc_loc}

\input{table/ago_loc}

\input{table/nusc_weather}

\noindent \textbf{Non-unique sorting.}
We show the difference between the random ordering strategy and an ordering based on coordinates in Table

\noindent \textbf{BEV augmentation.}
The first column of Table~\ref{tab:abl_aug} shows that the BEV augmentation provides a significant 2.7/3.3 improvement in both Landmark and Reachability scores.

\noindent \textbf{Synthetic noise objects.}
The second column of Table~\ref{tab:abl_aug} shows that the sequence augmentation of Synthetic noise objects technique~\cite{chen2021pix2seq}, however, leads to a drop in performance.
Whereas, the third column shows that the sequence noise padding 1.6/1.7 improved on both Landmark and Reachability scores. 
But the sequence noise padding is less effective than BEV augmentation.

\noindent \textbf{Class weight.}
We exam the class weight for MLE loss, \ie, $w$ for
\begin{equation}
    \max \sum_{i=1}^{L} w_i \log{P(\hat{y}_i|y_{<i}, \mathcal{F})}.
\end{equation}
\begin{equation}
    \max \sum_{i=1}^M \sum_{j=1}^L w_j P(y_{i, j}\ |\ \hat{y}, \mathcal{F}, \mathcal{V}_{kp}).
\end{equation}
Due to the high frequency of \texttt{Lineal} for $v_c$ and the default index for $v_d$, we assign a lower weight to these categories. Although the second column of Table~\ref{tab:abl_weight} does not indicate a clear relationship between class weight and performance, using a lower weight for the loss results in more stable performance.

\noindent \textbf{Embedding size.}
If we extend the embedding size from 576 to 1000 or 2000, useless embeddings clearly harm the performance.

\noindent \textbf{\strategy{}.} 
\ext{
In Table~\ref{tab:abl_tit}, we present the ablation study for \strategy{} (\stabb{}). In the first stage, we train using the lidar map as input. For the baseline, we use the same settings as \arabb{}, but change the input to the lidar map.
Since the input is simpler this time, adding more intense augmentation can significantly reduce overfitting. However, compared to \arabb{} in Table~\ref{tab:nus-reach-new}, the Landmark score shows a noticeable improvement due to the easier input modality. Nonetheless, the reachability performance is not as good as with the round-view image.
The potential problem is overfitting with this simple modality, so we use the sweep data, which is ten times the sample dataset. After training on the sweep data, both the landmark and reachability metrics achieve the best results.
}

\ext{
After we assemble LSS and \name{} in the third stage of \stabb{}, we study the difference between freezing and unfreezing the BEV Encoder. The difference is not significant, indicating that the distillation in stage 2 already aligns well with the training of \name{}. However, fine-tuning the BEV Encoder can slightly better match these two components.
}

\noindent \textbf{{Parameter fine-tuning.}}
{
We conducted experiments on hyperparameter tuning by varying the learning rate. The results, presented in Table~\ref{tab:re_nus-reach-lr}, show that our model performs best with the default setting and remains stable with minimal variation when the learning rate is adjusted.
}

\noindent \textbf{{Computational costs.}}
{
Besides the FPS reported in Table~\ref{tab:nus-reach}, we further provide details on the model parameters and inference time. 
The base model, AR-RNTR, has 37.25 million parameters, while NAR-RNTR has 35.90 million parameters and requires approximately 6GB of memory during inference with a batch size of 2, where the input images are 128 × 352.
}

\noindent \textbf{{Standard deviations.}}
{
Including standard deviations strengthens the reliability of our findings. We have provided the mean and standard deviation for \arabb{}, \sarabb{}, and \narabb{}, along with the number of trials. Each experiment was repeated five times, and the results are presented in Table~\ref{tab:re_nus-reach}. The performance of our models is statistically stable, with minimal variation.
}

\input{figure/mpr_curve}

\subsection{{Precision-Recall Curve}}
{
In addition to our overall advantage in mean Precision-Recall (as presented in Table 1 of the main submission), Figure~\ref{fig:mpr_curve} displays the precision/recall versus thresholds curve. Our models outperform others in terms of precision and recall for smaller thresholds, highlighting our accuracy advantage.
}

\subsection{{Robustness}}

{To evaluate the robustness of our method, we group the evaluation results by geographic region and report performance separately.
As shown in Table~\ref{tab:nusc-loc} and Table~\ref{tab:ago-loc}, our model consistently performs well across diverse urban environments. Specifically, in the nuScenes dataset, we observe strong and stable performance in both Singapore and Boston, demonstrating generalization across cities with different road structures and traffic styles. Similarly, on the Argoverse dataset, which contains data from Miami and Pittsburgh, our method also shows solid performance, indicating good adaptability to diverse North American cities.}

{Furthermore, to assess robustness under varying weather conditions, we conduct additional experiments on the nuScenes validation split by categorizing scenes based on annotated weather labels (i.e., normal, rainy, and cloudy). As reported in Table~\ref{tab:nusc-weather}, our method maintains strong performance across all weather conditions. This confirms the robustness and reliability of our model under challenging real-world scenarios.}

{Overall, the results across these three tables clearly demonstrate that our approach is resilient to variations in both geography and weather, supporting its effectiveness in real-world deployment.}

%% file: table/nus_stsu.tex
\begin{table}[tb]
  \centering
  \setlength{\tabcolsep}{0.5em}
  \caption{Comparison of front-camera \data{} extraction with state of the art on nuScenes~\cite{caesar2020nuscenes} PON validation split~\cite{roddick2020predicting}. 
    ResNet-50~\cite{he2016deep} is applied as image backbone by default.
    \texttt{M-P}, \texttt{M-R}, \texttt{M-F} stand for mean precision/recall/F1-score~\cite{can2021structured}.
    \texttt{Detect} stands for Detection ratio metrics~\cite{can2021structured}.
    \texttt{C-P}, \texttt{C-R}, \texttt{C-F} stand for connectivity precision/recall/F1-score~\cite{can2021structured}.
    }
    \begin{tabular}{l || C{0.7cm}C{0.7cm}C{0.7cm}|c|C{0.7cm}C{0.7cm}C{0.7cm}}
    \hline
    
    \hline
    Methods & M-P & M-R & M-F & Detect & C-P & C-R & C-F\\
    \hline

    PINET~\cite{ko2021key} & 54.1 & 45.6 & 49.5 & 19.2 & - & - & -\\
    Poly~\cite{acuna2018efficient} & 54.7 & 51.2 & 52.9 & 40.5 & 58.4 & 16.3 & 25.5\\
    STSU~\cite{can2021structured} & 60.7 & 54.7 & 57.5 & 60.6 & 60.5 & 52.2 & 56.0\\
    TPLR~\cite{can2022topology} & - & - & 58.2 & 60.2 & - & - & 55.3\\
    \hline
    \hline
    \arabb{} & 60.9 & 57.9 & 59.3 & 61.7 & 63.2 & 52.7 & 57.5\\
    \sarabb{} & \textbf{63.5} & \textbf{59.9} & \textbf{61.6} & \textbf{63.5} & \textbf{67.1} & \textbf{57.2} & \textbf{61.7}\\
    \narabb{} & 62.0 & 59.4 & 60.7 & 62.0 & 66.4 & 56.2 & 60.9\\
    \hline

    \hline
    \end{tabular}
  \label{tab:nus-stsu}
  \vspace{-4mm}
\end{table}

%% file: table/ago.tex
\begin{table}[tb]
  \centering

  \caption{
      Comparison on Argoverse~\cite{chang2019argoverse} dataset. ResNet-50~\cite{he2016deep} is applied as image backbone by default. M-F and C-F stand for mean and connectivity F1-score~\cite{can2021structured}. Detect stands for detection ratio metrics~\cite{can2021structured}.}
    
    \begin{tabular}{c||c|c|c}
    \hline
    
    \hline
    
    Methods & M-F & Detect & C-F \\
    \hline
    \hline
    
    PINET~\cite{ko2021key} & 47.2 & 15.1 & - \\
    TR~\cite{can2021structured} & 55.6 & 60.1 & 54.9 \\
    TPLR~\cite{can2022topology} & 57.1 & 64.2 & 58.1 \\
    Ours (AR-RNTR) & 57.6 & 64.5 & 58.2 \\

    \hline
    \hline

    \hline
    \end{tabular}
  \label{tab:re_ago}
\end{table}

%% file: table/nus_reach.tex
sc\begin{table}[tb]
  \centering
  \setlength{\tabcolsep}{0.2em}
  \caption{Comparison of multiple-camera \data{} extraction on nuScenes~\cite{caesar2020nuscenes} dataset assessed by Landmark Precision-Recall (\texttt{L-P}, \texttt{L-R}, \texttt{L-F}) and Reachability Precision-Recall (\texttt{R-P}, \texttt{R-R}, \texttt{R-F}).
    ResNet-50~\cite{he2016deep} is applied as image backbone by default.
    ``$\star$" use coupled \data{}. ``\dag" use VoVNetV2~\cite{lee2020centermask}
    }
    \begin{tabular}{l || C{0.7cm}C{0.7cm}C{0.7cm}|C{0.7cm}C{0.7cm}C{0.7cm}|C{1.6cm}}
    \hline
    
    \hline
    \multirow{2}{*}{Methods}& \multicolumn{3}{c|}{Landmark} & \multicolumn{3}{c|}{Reachability} & \multirow{2}{*}{FPS} \\
    \cline{2-7}
      & L-P & L-R & L-F & R-P & R-R & R-F &  \\
    \hline
    \hline
    \ext{\arabb{}}$\star$ & 60.9 & 46.6 & 52.8 & 73.2 & 49.7 & 59.2 & 0.1 (1.0$\times$)\\
    \ext{\arabb{}} & 62.5 & 47.7 & 54.1 & 73.2 & 52.8 & 61.3 & 0.1 (1.0$\times$)\\
    \ext{\sarabb{}} & \textbf{65.3} & \textbf{49.1} & \textbf{56.0} & \textbf{74.1} & \textbf{59.1} & \textbf{65.7} & 0.6 (6.0$\times$)\\
    \ext{\narabb{}} & 64.1 & 48.3 & 55.1 & 74.0 & 58.7 & 65.1 & \textbf{5.5} (47$\times$)\\
    \hline
    \arabb{}$\star$\dag & 62.6 & 47.9 & 54.3 & 73.2 & 52.9 & 61.4 & 0.1 (1.0$\times$)\\
    \sarabb{}\dag & \textbf{66.0} & \textbf{55.9} & \textbf{60.5} & \textbf{74.5} & \textbf{61.1} & \textbf{67.1} & 0.6 (6.0$\times$)\\
    \narabb{}\dag & 65.6 & 55.7 & 60.2 & 73.4 & 60.0 & 66.0 & \textbf{4.7} (47$\times$)\\
    \hline
    \ext{\arabb{} + \stabb{}} & 67.7 & 53.6 & 59.8 & 79.3 & 63.8 & 70.7 & 0.1 (1.0$\times$)\\

    \ext{\topoabb{} + \stabb{}} & \textbf{77.1} & \textbf{69.4} & \textbf{73.0} & \textbf{75.9} & \textbf{68.9} & \textbf{72.2} & 0.1 (1.0$\times$)\\
    \hline
    \hline

    \hline
    \end{tabular}
  \label{tab:nus-reach}
  \vspace{-4mm}
\end{table}

%% file: table/nus_reach_newsplit.tex
\begin{table}[tb]
  \centering
  \setlength{\tabcolsep}{0.3em}
  \caption{
    \ext{Comparison of multiple-camera \data{} extraction on nuScenes~\cite{caesar2020nuscenes} dataset with different split assessed by Landmark Precision-Recall (\texttt{L-P}, \texttt{L-R}, \texttt{L-F}) and Reachability Precision-Recall (\texttt{R-P}, \texttt{R-R}, \texttt{R-F}).
    ResNet-50~\cite{he2016deep} is applied as image backbone by default.``$\star$" use coupled \data{}.}
    }
    \begin{tabular}{l|l || C{0.7cm}C{0.7cm}C{0.7cm}|C{0.7cm}C{0.7cm}C{0.7cm}}
    \hline
    
    \hline
    \multirow{2}{*}{Split} & \multirow{2}{*}{Methods}& \multicolumn{3}{c|}{Landmark} & \multicolumn{3}{c}{Reachability}  \\
    \cline{3-8}
     &  & L-P & L-R & L-F & R-P & R-R & R-F  \\
     \hline
     \multirow{5}{*}{\ext{Default}} &\ext{\arabb{}}$\star$ & 60.9 & 46.6 & 52.8 & 73.2 & 49.7 & 59.2 \\
    &\ext{\arabb{}} & 62.5 & 47.7 & 54.1 & 73.2 & 52.8 & 61.3 \\
    &\ext{\sarabb{}} & \textbf{65.3} & \textbf{49.1} & \textbf{56.0} & \textbf{74.1} & \textbf{59.1} & \textbf{65.7} \\
    &\ext{\narabb{}} & 64.1 & 48.3 & 55.1 & 74.0 & 58.7 & 65.1 \\
    \cline{2-8}
    &\ext{\arabb{} + \stabb{}} & 67.7 & 53.6 & 59.8 & 79.3 & 63.8 & 70.7\\

    &\ext{\topoabb{} + \stabb{}} & \textbf{77.1 }& \textbf{69.4} & \textbf{73.0} & \textbf{75.9} & \textbf{68.9} & \textbf{72.2}\\
    \hline
    \hline
    \multirow{5}{*}{\ext{New}} & \ext{\arabb{}} & 47.7 & \textbf{29.6} & 36.5 & 52.1 & 33.6 & 40.9 \\
     &\ext{\sarabb{}} & \textbf{50.3} & 29.2 & \textbf{36.9} & \textbf{55.9} & \textbf{35.0} & \textbf{43.0} \\
    &\ext{\narabb{}} & 48.7 & 28.8 & 36.2 & 55.6 & 34.1 & 42.3 \\
    \cline{2-8}
    &\ext{\arabb{} + \stabb{}} & 48.8 & 32.7 & 39.1 & 67.1 & 39.6 & 49.8\\

    &\ext{\topoabb{} + \stabb{}} & \textbf{63.0}& \textbf{48.3} & \textbf{54.7} & \textbf{67.9} & \textbf{58.1} & \textbf{62.6}\\
    \hline

    \hline
    \end{tabular}
  \label{tab:nus-reach-new}
  \vspace{-6mm}
\end{table}

%% file: table/ablation_layer.tex
\begin{table}[htb]
  \centering
  \setlength{\tabcolsep}{0.2em}
  \caption{Ablation studies on number of Transformer decoder layers and Intra-seq self-attention of \narabb{}. \texttt{\# Key-pt} denotes Key-point Transformer Decoder layer number, and \texttt{\# Para-seq} denotes Parallel-seq Transformer Decoder layer number. \texttt{Intra-Seq} means Intra-seq self-attention in Parallel-seq Transformer Decoder layer.
    VoVNetV2~\cite{lee2020centermask} pretrained on extra data is applied as image backbone by default.
    The row with gray color is our final choice.
    }
    \begin{tabular}{ C{1.6cm}C{1.6cm}C{1.6cm}||C{0.7cm}C{0.7cm}|C{1.6cm}}
    \hline
    
    \hline
    \# Key-pt & \# Para-Seq & Intra-Seq & L-F & R-F & FPS\\
    
    \hline
     3 & 3 & \checkmark & 56.1 & 63.3 & \textbf{4.8} \\
     5 & 3 & \checkmark & 58.2 & 65.1 & 4.8 \\
     \rowcolor[gray]{.9} 
     6 & 3 & \checkmark & 60.2 & 66.0 & 4.7 \\
     6 & 5 & \checkmark & 60.5 & 66.9 & 3.4 \\
     6 & 6 & \checkmark & \textbf{61.3} & \textbf{67.1} & 3.0 \\
     6 & 3 & \XSolidBrush & 55.7 & 56.3 & 5.0 \\
    \hline

    \hline
    \end{tabular}
  \label{tab:abl_layers}
  \vspace{-4mm}
\end{table}

%% file: table/ablation_mask_iter.tex
\begin{table}[htb]
  \centering
  \setlength{\tabcolsep}{0.6em}
  \caption{Ablation studies on mask ratio and iteration times of \narabb{}. 
    VoVNetV2~\cite{lee2020centermask} pretrained on extra data is applied as image backbone by default.
    The row with gray color is our final choice.
    }
    \begin{tabular}{ C{1.9cm}C{1.9cm}||C{0.7cm}C{0.7cm}|C{1.6cm}}
    \hline
    
    \hline
    Mask ratio & \# iteration & L-F & R-F & FPS\\
    
    \hline
    50\% & 3 & 51.4 & 51.3 & 4.7 \\
    75\% & 3 & 59.4 & 62.7 & 4.7 \\
     90\% & 1 & 59.0 & 62.7 & \textbf{8.9} \\
     \rowcolor[gray]{.9} 
     90\% & 3 & 60.2 & 66.0 & 4.7 \\
     90\% & 6 & \textbf{60.3} & \textbf{66.1} & 2.8  \\
    \hline

    \hline
    \end{tabular}
  \label{tab:abl_mask_iter}
  \vspace{-2mm}
\end{table}

%% file: table/ablation_aug.tex
\begin{table}[tb]
  \centering
  \setlength{\tabcolsep}{0.2em}
  \caption{Ablation studies on BEV augmentation and synthetic noise objects~\cite{chen2021pix2seq} (including sequence augmentation and sequence noise padding).
    \narabb{} with VoVNetV2~\cite{lee2020centermask} pretrained on extra data is applied.
    The row with gray color is our final choice.
    }
    \begin{tabular}{ C{1.9cm}C{2.2cm}C{2.4cm}||C{0.7cm}C{0.7cm}}
    \hline
    
    \hline
    BEV Aug & Sequence Aug & Sequence Noise & L-F & R-F \\
    \hline
    \checkmark & \XSolidBrush & \XSolidBrush & 58.6 & 64.3\\
    \XSolidBrush & \XSolidBrush & \checkmark & 57.5 & 62.7\\
    \rowcolor[gray]{.9} 
    \checkmark & \XSolidBrush & \checkmark & 60.2 & 66.0\\
    \checkmark & \checkmark & \checkmark & 59.1 & 65.2\\
    \hline

    \hline
    \end{tabular}
    
  \label{tab:abl_aug}
\end{table}

%% file: table/ablation_weight.tex
\begin{table}[tb]
  \centering
  \caption{Ablation studies on sequence embedding size and class weight.
    \narabb{} with VoVNetV2~\cite{lee2020centermask} pretrained on extra data is applied.
    The row with gray color is our final choice.
    }
    \begin{tabular}{ C{2.4cm}C{1.9cm}||C{0.7cm}C{0.7cm}}
    \hline
    
    \hline
    Embedding size & class weight & L-F & R-F \\
    \hline
    576 & 1.0 & 60.1 & 65.5\\
    576 & 0.5 & 60.1 & \textbf{66.1}\\
    576 & 0.1 & 60.2 & 66.0\\
    \hline
    \rowcolor[gray]{.9}
    576 & 0.2 & \textbf{60.2} & 66.0\\
    \hline
    1000 & 0.2 & 60.1 & 65.8\\
    2000 & 0.2 & 59.7 & 65.5\\
    \hline

    \hline
    \end{tabular}
  \label{tab:abl_weight}
\end{table}

%% file: table/ablation_tit.tex
\begin{table}[htb]
  \centering
  \setlength{\tabcolsep}{0.3em}
  \caption{
    \ext{
Ablation studies on \strategy{}. Models with ResNet~\cite{he2016deep} trained on the nuScenes~\cite{caesar2020nuscenes} non-overfitting dataset are used. The "Lidar Map" stage indicates the first stage of \stabb{}, and "Assembled" represents the third stage of \stabb{}. "Sweep data" uses 20Hz sampling rate data from the nuScenes dataset.
    }
    }
    \begin{tabular}{l|l || C{0.7cm}C{0.7cm}C{0.7cm}|C{0.7cm}C{0.7cm}C{0.7cm}}
    \hline
    
    \hline
    \multirow{2}{*}{Stage} & \multirow{2}{*}{Methods}& \multicolumn{3}{c|}{Landmark} & \multicolumn{3}{c}{Reachability}  \\
    \cline{3-8}
     & & L-P & L-R & L-F & R-P & R-R & R-F  \\
     \hline
     \multirow{3}{*}{\ext{Lidar Map}}  & \ext{Baseline}  & 47.8 & 38.9 & 42.9 & 60.4 & 24.9 & 35.3\\
     &\ext{ + More aug} & 52.6 & 41.4 & 46.3 & 70.9 & 25.6 & 37.6\\
     &\ext{ + Sweep data} & \textbf{67.2} & \textbf{47.2} & \textbf{55.4} & \textbf{81.1} & \textbf{51.0} & \textbf{62.6}\\
     \hline
     \multirow{2}{*}{\ext{Assembled}} & \ext{Freeze BEV} & 47.7 & \textbf{34.2} & 38.6 & 66.7 & 38.2 & 48.6\\
     & \ext{Unfreeze BEV} & \textbf{48.8} & 32.7 & \textbf{39.1} & \textbf{67.1} &\textbf{39.6} & \textbf{49.8}\\
    \hline

    \hline
    \end{tabular}
  \label{tab:abl_tit}
\end{table}

%% file: table/param_finetune.tex
\begin{table}[tb]
  \centering

  \caption{
      {Comparison on nuScenes~\cite{caesar2020nuscenes} dataset assessed by Landmark Precision-Recall (\texttt{L-P}, \texttt{L-R}, \texttt{L-F}) and Reachability Precision-Recall (\texttt{R-P}, \texttt{R-R}, \texttt{R-F}). The model used is \arabb{}. ResNet-50~\cite{he2016deep} is applied as image backbone by default.}
    }
    
    \begin{tabular}{c||ccc|ccc}
    \hline
    
    \hline
    \hline
    
    \multirow{2}{*}{Learning Rate}& \multicolumn{3}{c|}{Landmark} & \multicolumn{3}{c}{Reachability} \\
    \cline{2-7}
      & L-P & L-R & L-F & R-P & R-R & R-F \\
    \hline
    \hline
    
    {1e-4} & {61.0} & {46.8} & {53.0} & {72.5} & {52.1} & {60.6} \\
    {2e-4} & \textbf{{62.5}} & \textbf{{47.7}} & \textbf{{54.1}} & \textbf{{73.2}} & \textbf{{52.8}} & \textbf{{61.3}} \\
    {3e-4} & {62.0} & {47.2} & {54.0} & {72.8} & {52.4} & {60.7} \\

    \hline
    \hline

    \hline
    \end{tabular}
  \label{tab:re_nus-reach-lr}
\end{table}

%% file: table/standard_deviation.tex
\begin{table}[ht!]
  \centering
  \setlength{\tabcolsep}{0.2em}
  \caption{
      {Comparison on nuScenes~\cite{caesar2020nuscenes} dataset assessed by Landmark Precision-Recall (\texttt{L-P}, \texttt{L-R}, \texttt{L-F}) and Reachability Precision-Recall (\texttt{R-P}, \texttt{R-R}, \texttt{R-F}). ResNet-50~\cite{he2016deep} is applied as image backbone by default.}
    }
    
    \begin{tabular}{l||ccc|ccc}
    \hline
    
    \hline
    \hline
    
    \multirow{2}{*}{Methods}& \multicolumn{3}{c|}{Landmark} & \multicolumn{3}{c}{Reachability} \\
    \cline{2-7}
      & L-P & L-R & L-F & R-P & R-R & R-F \\
    \hline
    \hline
    
    {\arabb{}} & {62.5$\pm$0.2} & {47.7$\pm$0.1} & {54.1$\pm$0.1} & {73.2$\pm$0.1} & {52.8$\pm$0.2} & {61.3$\pm$0.1} \\
    {\sarabb{}} & \textbf{{65.3$\pm$0.1}} & \textbf{{49.1$\pm$0.1}} & \textbf{{56.0$\pm$0.1}} & \textbf{{74.1$\pm$0.1}} & \textbf{{59.1$\pm$0.2}} & \textbf{{65.7$\pm$0.2}} \\
    {\narabb{}} & {64.1$\pm$0.1} & {48.3$\pm$0.1} & {55.1$\pm$0.2} & {74.0$\pm$0.1} & {58.7$\pm$0.3} & {65.1$\pm$0.1} \\

    \hline
    \hline

    \hline
    \end{tabular}
  \label{tab:re_nus-reach}
\end{table}

%% file: table/nusc_loc.tex
\begin{table*}[ht!]
  \centering
  \caption{{Comparison of different geographical location settings on nuScenes~\cite{caesar2020nuscenes} PON validation split~\cite{roddick2020predicting}. 
    ResNet-50~\cite{he2016deep} is applied as image backbone by default.
    \texttt{M-P}, \texttt{M-R}, \texttt{M-F} stand for mean precision/recall/F1-score~\cite{can2021structured}.
    \texttt{Detect} stands for Detection ratio metrics~\cite{can2021structured}.
    \texttt{C-P}, \texttt{C-R}, \texttt{C-F} stand for connectivity precision/recall/F1-score~\cite{can2021structured}.}
    }
    \begin{tabular}{l || ccccccc | ccccccc }
    \hline
    \hline
    \multirow{3}{*}{{Method}} & \multicolumn{14}{c}{{nuScenes}} \\
    \cline{2-15}
    & \multicolumn{7}{c|}{{Singapore}} & \multicolumn{7}{c}{{Boston}} \\
    & {M-P} & {M-R} & {M-F} & {Detect} & {C-P} & {C-R} & {C-F} & {M-P} & {M-R} & {M-F} & {Detect} & {C-P} & {C-R} & {C-F} \\
    \hline
    \hline
    {\arabb{}}   & {61.1} & {57.8} & {59.6} & {61.8} & {63.1} & {52.5} & {57.9} & {60.8} & {58.2} & {59.1} & {61.5} & {63.4} & {52.8} & {57.3} \\
    {\sarabb{}}  & {63.8} & {60.4} & {62.0} & {63.4} & {67.3} & {56.8} & {61.2} & {63.0} & {59.6} & {61.1} & {63.5} & {66.8} & {57.5} & {62.0} \\
    {\narabb{}}  & {61.4} & {59.1} & {60.2} & {62.3} & {66.8} & {55.7} & {60.5} & {62.5} & {59.6} & {61.1} & {61.8} & {66.2} & {56.4} & {61.2} \\
    \hline
    \hline
    \end{tabular}
  \label{tab:nusc-loc}
\end{table*}

%% file: table/ago_loc.tex
\begin{table}[ht!]
  \centering
  \caption{{Comparison of different geographical location settings on Argoverse~\cite{chang2019argoverse} dataset. ResNet-50~\cite{he2016deep} is applied as image backbone by default. M-F and C-F stand for mean and connectivity F1-score~\cite{can2021structured}. Detect stands for detection ratio metrics~\cite{can2021structured}.}
  }
    \begin{tabular}{l || ccc | ccc}
    \hline
    \hline
    \multirow{3}{*}{{Method}} & \multicolumn{6}{c}{{Argoverse}} \\
    \cline{2-7}
    & \multicolumn{3}{c|}{{Miami}} & \multicolumn{3}{c}{{Pittsburgh}} \\
    & {M-F} & {Detect} & {C-F} & {M-F} & {Detect} & {C-F} \\
    \hline
    \hline
    {\arabb{}}   & {58.2} & {64.8} & {58.4} & {57.2} & {64.1} & {58.1} \\
    {\sarabb{}}  & {60.1} & {65.4} & {61.2} & {60.6} & {66.3} & {62.0} \\
    {\narabb{}}  & {58.4} & {65.6} & {60.7} & {57.6} & {64.3} & {60.2} \\
    \hline
    \hline
    \end{tabular}
  \label{tab:ago-loc}
\end{table}

%% file: table/nusc_weather.tex
\begin{table}[ht!]
  \centering
  \setlength{\tabcolsep}{0.3em}
  \caption{{Comparison of different weather condition settings on nuScenes~\cite{caesar2020nuscenes} PON validation split~\cite{roddick2020predicting}. 
    ResNet-50~\cite{he2016deep} is applied as image backbone by default.
    \texttt{M-F} stand for mean F1-score~\cite{can2021structured}.
    \texttt{Detect} stands for Detection ratio metrics~\cite{can2021structured}.
    \texttt{C-F} stand for connectivity F1-score~\cite{can2021structured}.}
    }
    \begin{tabular}{l || ccc | ccc | ccc }
    \hline
    \hline
    \multirow{3}{*}{{Method}} & \multicolumn{9}{c}{{nuScenes}} \\
    \cline{2-10}
    & \multicolumn{3}{c|}{{Normal}} & \multicolumn{3}{c|}{{Rainy}} & \multicolumn{3}{c}{{Cloudy}} \\
    & {M-F} & {Detect} & {C-F} & {M-F} & {Detect} & {C-F} & {M-F} & {Detect} & {C-F} \\
    \hline
    \hline
    {\arabb{}}   & {59.9} & {62.8} & {58.4} & {58.6} & {61.3} & {56.9} & {59.0} & {61.5} & {57.3} \\
    {\sarabb{}}  & {62.4} & {63.9} & {62.6} & {60.7} & {63.1} & {61.1} & {61.8} & {63.6} & {62.0} \\
    {\narabb{}}  & {62.0} & {63.5} & {62.1} & {60.3} & {61.5} & {60.3} & {60.2} & {61.7} & {60.4} \\ 
    \hline
    \hline
    \end{tabular}
  \label{tab:nusc-weather}
\end{table}

%% file: figure/mpr_curve.tex
\begin{figure}[htb]
	\def \imwidth {4.3cm}
	\def \imheight {3.2cm}
	\centering
	\includegraphics[height=\imheight, width=\imwidth]{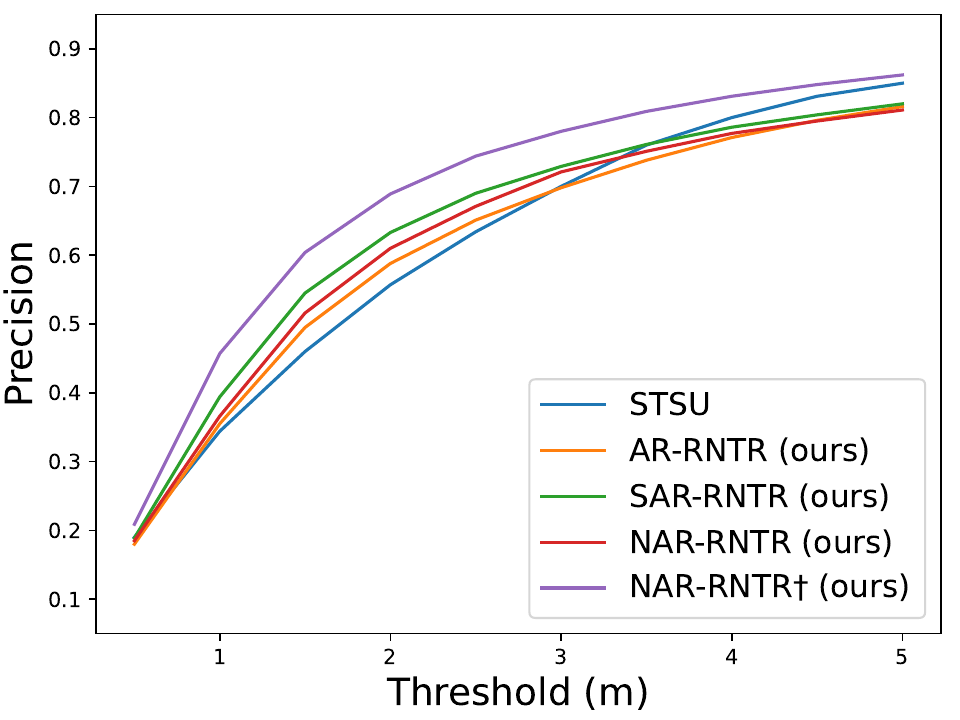}
	\includegraphics[height=\imheight, width=\imwidth]{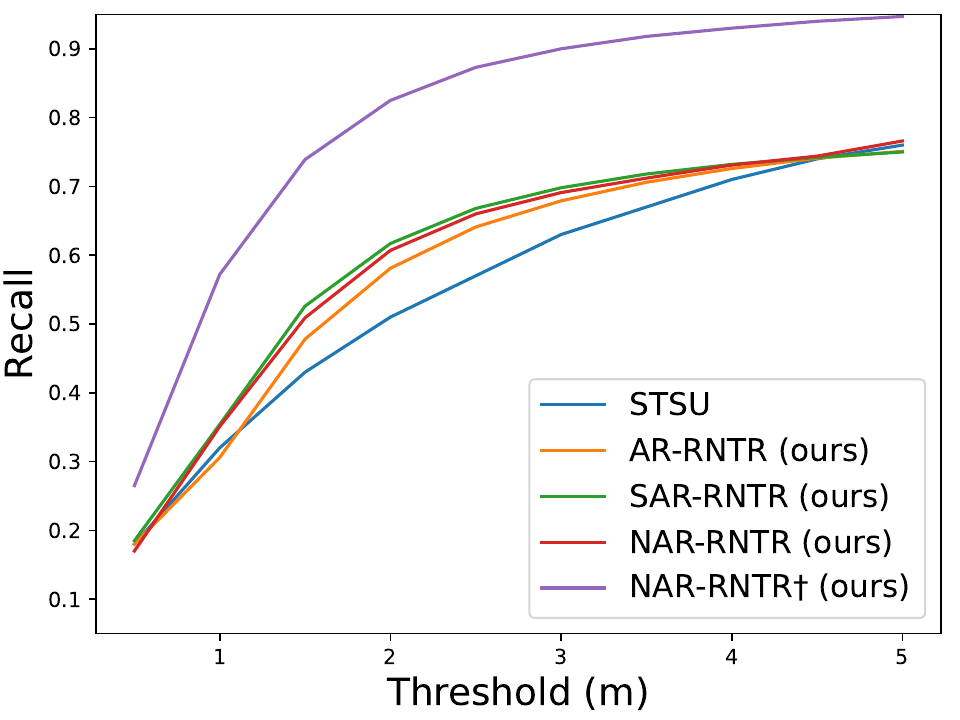}\\

	\caption{{Mean Precision/Recall v.s. thresholds. ``$\dag$" use VoVNetV2~\cite{lee2020centermask} pretrained on extra data as backbone.
    Thresholds are from $[0.5, 1.0, 1.5, 2.0, 2.5, 3.0, 3.5, 4.0, 4.5, 5.0]m$.}
	}
	\label{fig:mpr_curve}
    \vspace{-4mm}
\end{figure}

%% file: file/5-conclusion.tex
\section{Conclusion and Discussions}
\noindent\textbf{Conclusion.}
In summary, our work introduces a lossless, efficient, and interactional sequence representation called \seqdata{}, which preserves both Euclidean and non-Euclidean data of road networks.
We have designed an Auto-Regressive \name{} as a baseline model, which takes advantage of the auto-regressive dependency of \seqdata{}. 
Additionally, we have proposed Semi-Autoregressive and Non-Autoregressive \name{} models, which decouple the auto-regressive dependency of \seqdata{}, resulting in significantly faster inference speeds and improved accuracy. 
Our extensive experiments demonstrate the superiority of our \seqdata{} representation and \name{} models.

\ext{
To further adapt \name{} for real-world applications, we conduct more experiments on the non-overfitting split of the nuScenes~\cite{caesar2020nuscenes} dataset. We identify two main bottlenecks in the current \name{}: poor landmark detection limited by the BEV Encoder and error propagation to topology reasoning. Therefore, we propose \strategy{} to inherit better topology knowledge into \name{}.
Additionally, we collect SD-Maps from open-source map datasets and use this prior information to significantly improve landmark detection and reachability.
}

\noindent\textbf{{Limitations.}}
{While our method demonstrates strong performance across various metrics and datasets, there are still some limitations that deserve future exploration.
First, our current BEV Encoder relies on 2D projections, which may struggle to capture elevation or fine-grained vertical structures (e.g., tunnels, flyovers), especially in complex 3D scenes.
Second, while SD-Maps offer valuable priors, their coarse resolution limits their effectiveness for precise landmark localization, which may restrict performance in high-precision navigation scenarios.
We believe these challenges provide promising directions for future work, including exploring 3D aware representations, and leveraging richer map priors with higher spatial fidelity.}

%% file: file/6-appendix-new.tex
\section{supplemental material}

\input{figure/visualization1}

\input{figure/visualization_pretrain_sd}

\subsection{Qualitative Results}
We present the visualization on the overfitting dataset in Figure~\ref{fig:visualization}.
The precise localization of landmarks, accurate topological connections, and precise curve shapes demonstrate the superiority of \name{}.

\ext{
We also compare \arabb{}, \strategy{}, and SD-Maps on the nuScenes~\cite{caesar2020nuscenes} non-overfitting split.
As shown in Figure~\ref{fig:visualization_pretrain_sd}, after adding \strategy{}, the landmark predictions are more precise, and the connections are reasonable in most cases. With the addition of SD-Maps, the results finally become practical for real applications.
}

%% file: figure/visualization1.tex
\begin{figure*}[b]
    \centering
    \includegraphics[width=\linewidth]{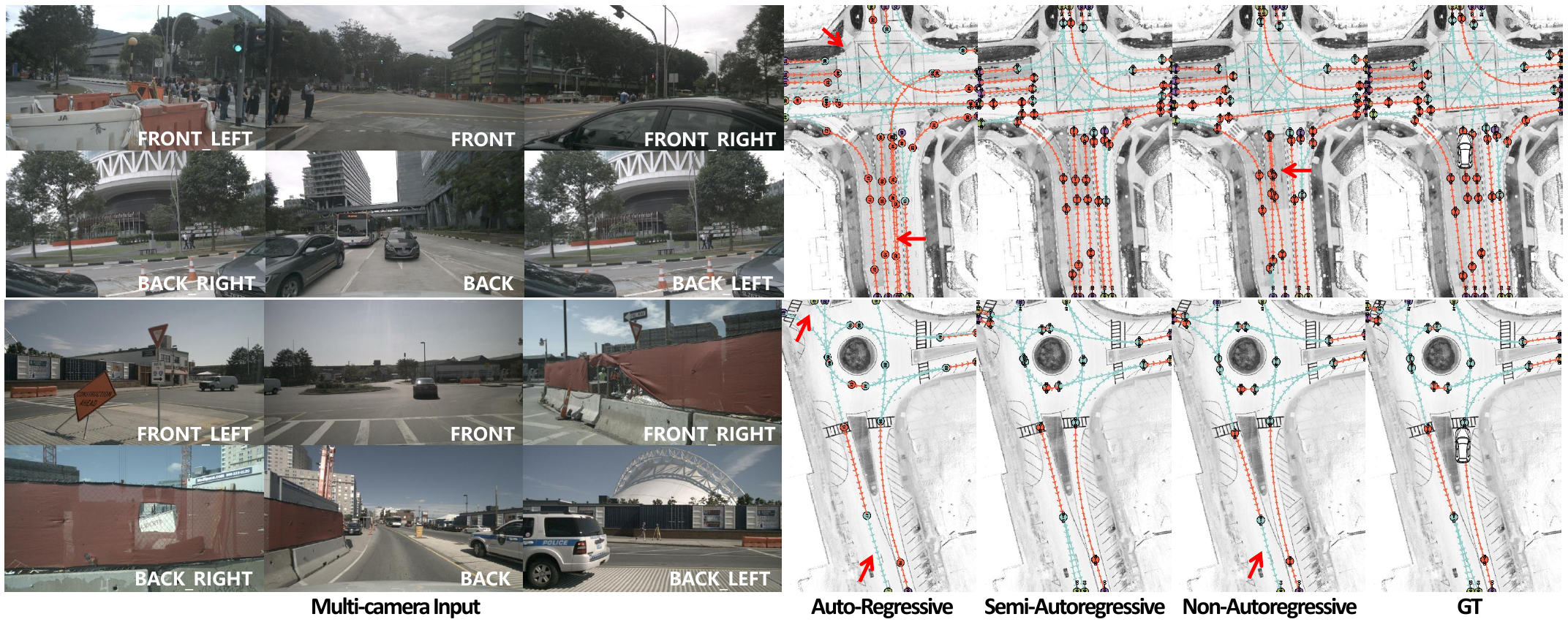}
    \caption{Qualitative results on nuScenes validation set.
    All three variants of \name{} predict high quality road network.
    Only slight errors (pointed by red arrow) occur when predicting landmarks locations for \arabb{} and \narabb{}.
    }
    \label{fig:visualization}
    \vspace{-4mm}
\end{figure*}

%% file: figure/visualization_pretrain_sd.tex
\begin{figure*}[b]
    \centering
    \includegraphics[width=\linewidth]{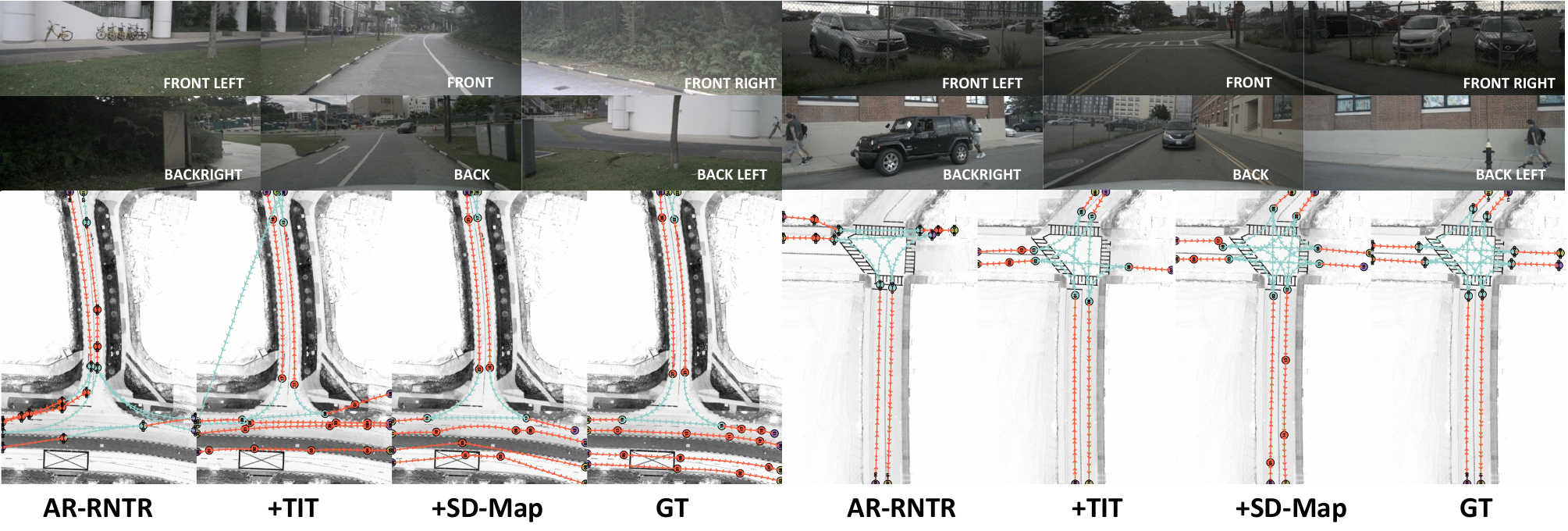}
    \caption{
    \ext{Qualitative results on the nuScenes~\cite{caesar2020nuscenes} non-overfitting validation set. Both \arabb{} and \arabb{} with \strategy{} do not perform ideally in the non-overfitting case. However, with the integration of SD-Maps, \arabb{} can finally produce high-quality and usable \data{}.}
    }
    \label{fig:visualization_pretrain_sd}
    \vspace{-4mm}
\end{figure*}

%% file: main_pure.bbl
\begin{thebibliography}{10}
\providecommand{\url}[1]{#1}
\csname url@samestyle\endcsname
\providecommand{\newblock}{\relax}
\providecommand{\bibinfo}[2]{#2}
\providecommand{\BIBentrySTDinterwordspacing}{\spaceskip=0pt\relax}
\providecommand{\BIBentryALTinterwordstretchfactor}{4}
\providecommand{\BIBentryALTinterwordspacing}{\spaceskip=\fontdimen2\font plus
\BIBentryALTinterwordstretchfactor\fontdimen3\font minus \fontdimen4\font\relax}
\providecommand{\BIBforeignlanguage}[2]{{%
\expandafter\ifx\csname l@#1\endcsname\relax
\typeout{** WARNING: IEEEtran.bst: No hyphenation pattern has been}%
\typeout{** loaded for the language `#1'. Using the pattern for}%
\typeout{** the default language instead.}%
\else
\language=\csname l@#1\endcsname
\fi
#2}}
\providecommand{\BIBdecl}{\relax}
\BIBdecl

\bibitem{cui2019multimodal}
H.~Cui, V.~Radosavljevic, F.-C. Chou, T.-H. Lin, T.~Nguyen, T.-K. Huang, J.~Schneider, and N.~Djuric, ``Multimodal trajectory predictions for autonomous driving using deep convolutional networks,'' in \emph{IEEE International Conference on Robotics and Automation}, 2019.

\bibitem{hong2019rules}
J.~Hong, B.~Sapp, and J.~Philbin, ``Rules of the road: Predicting driving behavior with a convolutional model of semantic interactions,'' in \emph{IEEE Conference on Computer Vision and Pattern Recognition}, 2019.

\bibitem{rella2021decoder}
E.~M. Rella, J.-N. Zaech, A.~Liniger, and L.~Van~Gool, ``Decoder fusion rnn: Context and interaction aware decoders for trajectory prediction,'' in \emph{International Conference on Intelligent Robots and Systems}, 2021.

\bibitem{chen2020learning}
D.~Chen, B.~Zhou, V.~Koltun, and P.~Kr{\"a}henb{\"u}hl, ``Learning by cheating,'' in \emph{Conference on Robot Learning}, 2020.

\bibitem{casas2021mp3}
S.~Casas, A.~Sadat, and R.~Urtasun, ``Mp3: A unified model to map, perceive, predict and plan,'' in \emph{IEEE Conference on Computer Vision and Pattern Recognition}, 2021.

\bibitem{ma2019exploiting}
W.-C. Ma, I.~Tartavull, I.~A. B{\^a}rsan, S.~Wang, M.~Bai, G.~Mattyus, N.~Homayounfar, S.~K. Lakshmikanth, A.~Pokrovsky, and R.~Urtasun, ``Exploiting sparse semantic hd maps for self-driving vehicle localization,'' in \emph{International Conference on Intelligent Robots and Systems}, 2019.

\bibitem{ravi2018real}
B.~Ravi~Kiran, L.~Roldao, B.~Irastorza, R.~Verastegui, S.~Suss, S.~Yogamani, V.~Talpaert, A.~Lepoutre, and G.~Trehard, ``Real-time dynamic object detection for autonomous driving using prior 3d-maps,'' in \emph{European Conference on Computer Vision workshops}, 2018.

\bibitem{bronstein2017geometric}
M.~M. Bronstein, J.~Bruna, Y.~LeCun, A.~Szlam, and P.~Vandergheynst, ``Geometric deep learning: going beyond euclidean data,'' \emph{IEEE Signal Processing Magazine}, vol.~34, no.~4, pp. 18--42, 2017.

\bibitem{bronstein2021geometric}
M.~M. Bronstein, J.~Bruna, T.~Cohen, and P.~Veli{\v{c}}kovi{\'c}, ``Geometric deep learning: Grids, groups, graphs, geodesics, and gauges,'' \emph{arXiv preprint}, 2021.

\bibitem{xu2020curvelane}
H.~Xu, S.~Wang, X.~Cai, W.~Zhang, X.~Liang, and Z.~Li, ``Curvelane-nas: Unifying lane-sensitive architecture search and adaptive point blending,'' in \emph{European Conference on Computer Vision}, 2020.

\bibitem{liu2021condlanenet}
L.~Liu, X.~Chen, S.~Zhu, and P.~Tan, ``Condlanenet: a top-to-down lane detection framework based on conditional convolution,'' in \emph{IEEE International Conference on Computer Vision}, 2021.

\bibitem{xu2022rclane}
S.~Xu, X.~Cai, B.~Zhao, L.~Zhang, H.~Xu, Y.~Fu, and X.~Xue, ``Rclane: Relay chain prediction for lane detection,'' in \emph{European Conference on Computer Vision}, 2022.

\bibitem{philion2020lift}
J.~Philion and S.~Fidler, ``Lift, splat, shoot: Encoding images from arbitrary camera rigs by implicitly unprojecting to 3d,'' in \emph{European Conference on Computer Vision}, 2020.

\bibitem{saha2022translating}
A.~Saha, O.~Mendez, C.~Russell, and R.~Bowden, ``Translating images into maps,'' in \emph{IEEE International Conference on Robotics and Automation}, 2022.

\bibitem{li2022bevformer}
Z.~Li, W.~Wang, H.~Li, E.~Xie, C.~Sima, T.~Lu, Y.~Qiao, and J.~Dai, ``Bevformer: Learning bird’s-eye-view representation from multi-camera images via spatiotemporal transformers,'' in \emph{European Conference on Computer Vision}, 2022.

\bibitem{lu2022learning}
J.~Lu, Z.~Zhou, X.~Zhu, H.~Xu, and L.~Zhang, ``Learning ego 3d representation as ray tracing,'' in \emph{European Conference on Computer Vision}, 2022.

\bibitem{can2021structured}
Y.~B. Can, A.~Liniger, D.~P. Paudel, and L.~Van~Gool, ``Structured bird's-eye-view traffic scene understanding from onboard images,'' in \emph{IEEE Conference on Computer Vision and Pattern Recognition}, 2021.

\bibitem{can2022topology}
Y.~B. \vspace{0mm}Can, A.~Liniger, D.~P. Paudel, and L.~Van~Gool, ``Topology preserving local road network estimation from single onboard camera image,'' in \emph{IEEE Conference on Computer Vision and Pattern Recognition}, 2022.

\bibitem{acuna2018efficient}
D.~Acuna, H.~Ling, A.~Kar, and S.~Fidler, ``Efficient interactive annotation of segmentation datasets with polygon-rnn++,'' in \emph{IEEE Conference on Computer Vision and Pattern Recognition}, 2018.

\bibitem{vaswani2017attention}
A.~Vaswani, N.~Shazeer, N.~Parmar, J.~Uszkoreit, L.~Jones, A.~N. Gomez, {\L}.~Kaiser, and I.~Polosukhin, ``Attention is all you need,'' in \emph{Advances in Neural Information Processing Systems}, 2017.

\bibitem{brown2020language}
T.~Brown, B.~Mann, N.~Ryder, M.~Subbiah, J.~D. Kaplan, P.~Dhariwal, A.~Neelakantan, P.~Shyam, G.~Sastry, A.~Askell \emph{et~al.}, ``Language models are few-shot learners,'' in \emph{Advances in Neural Information Processing Systems}, 2020.

\bibitem{chen2021pix2seq}
T.~Chen, S.~Saxena, L.~Li, D.~J. Fleet, and G.~Hinton, ``Pix2seq: A language modeling framework for object detection,'' in \emph{International Conference on Learning Representations}, 2021.

\bibitem{lee2018deterministic}
J.~Lee, E.~Mansimov, and K.~Cho, ``Deterministic non-autoregressive neural sequence modeling by iterative refinement,'' in \emph{Conference on Empirical Methods in Natural Language Processing}, 2018.

\bibitem{caesar2020nuscenes}
H.~Caesar, V.~Bankiti, A.~H. Lang, S.~Vora, V.~E. Liong, Q.~Xu, A.~Krishnan, Y.~Pan, G.~Baldan, and O.~Beijbom, ``nuscenes: A multimodal dataset for autonomous driving,'' in \emph{IEEE Conference on Computer Vision and Pattern Recognition}, 2020.

\bibitem{roddick2020predicting}
T.~Roddick and R.~Cipolla, ``Predicting semantic map representations from images using pyramid occupancy networks,'' in \emph{IEEE Conference on Computer Vision and Pattern Recognition}, 2020.

\bibitem{googlemaps}
{Google}, ``Google maps,'' \url{https://maps.google.com}, accessed: 2024-06-16.

\bibitem{openstreetmap}
{OpenStreetMap contributors}, ``Openstreetmap,'' \url{https://www.openstreetmap.org}, accessed: 2024-06-16.

\bibitem{roddick2018orthographic}
T.~Roddick, A.~Kendall, and R.~Cipolla, ``Orthographic feature transform for monocular 3d object detection,'' in \emph{British Machine Vision Conference}, 2019.

\bibitem{huang2021bevdet}
J.~Huang, G.~Huang, Z.~Zhu, and D.~Du, ``Bevdet: High-performance multi-camera 3d object detection in bird-eye-view,'' \emph{arXiv preprint}, 2021.

\bibitem{hu2021fiery}
A.~Hu, Z.~Murez, N.~Mohan, S.~Dudas, J.~Hawke, V.~Badrinarayanan, R.~Cipolla, and A.~Kendall, ``Fiery: future instance prediction in bird's-eye view from surround monocular cameras,'' in \emph{IEEE International Conference on Computer Vision}, 2021.

\bibitem{zhou2023suit}
Z.~Zhou, J.~Lu, Y.~Zeng, H.~Xu, and L.~Zhang, ``Suit: Learning significance-guided information for 3d temporal detection,'' \emph{arXiv preprint arXiv:2307.01807}, 2023.

\bibitem{gao2020vectornet}
J.~Gao, C.~Sun, H.~Zhao, Y.~Shen, D.~Anguelov, C.~Li, and C.~Schmid, ``Vectornet: Encoding hd maps and agent dynamics from vectorized representation,'' in \emph{IEEE Conference on Computer Vision and Pattern Recognition}, 2020.

\bibitem{liao2022maptr}
B.~Liao, S.~Chen, X.~Wang, T.~Cheng, Q.~Zhang, W.~Liu, and C.~Huang, ``Maptr: Structured modeling and learning for online vectorized hd map construction,'' \emph{arXiv preprint}, 2022.

\bibitem{wang2019pseudo}
Y.~Wang, W.-L. Chao, D.~Garg, B.~Hariharan, M.~Campbell, and K.~Q. Weinberger, ``Pseudo-lidar from visual depth estimation: Bridging the gap in 3d object detection for autonomous driving,'' in \emph{IEEE Conference on Computer Vision and Pattern Recognition}, 2019.

\bibitem{reading2021categorical}
C.~Reading, A.~Harakeh, J.~Chae, and S.~L. Waslander, ``Categorical depth distribution network for monocular 3d object detection,'' in \emph{IEEE Conference on Computer Vision and Pattern Recognition}, 2021.

\bibitem{ruan2020learning}
S.~Ruan, C.~Long, J.~Bao, C.~Li, Z.~Yu, R.~Li, Y.~Liang, T.~He, and Y.~Zheng, ``Learning to generate maps from trajectories,'' in \emph{AAAI Conference on Artificial Intelligence}, 2020.

\bibitem{wu2020deepdualmapper}
H.~Wu, H.~Zhang, X.~Zhang, W.~Sun, B.~Zheng, and Y.~Jiang, ``Deepdualmapper: A gated fusion network for automatic map extraction using aerial images and trajectories,'' in \emph{AAAI Conference on Artificial Intelligence}, 2020.

\bibitem{gu2017non}
J.~Gu, J.~Bradbury, C.~Xiong, V.~O. Li, and R.~Socher, ``Non-autoregressive neural machine translation,'' in \emph{International Conference on Learning Representations}, 2018.

\bibitem{kim2016sequence}
Y.~Kim and A.~M. Rush, ``Sequence-level knowledge distillation,'' in \emph{Conference on Empirical Methods in Natural Language Processing}, 2016.

\bibitem{zhou2019understanding}
C.~Zhou, G.~Neubig, and J.~Gu, ``Understanding knowledge distillation in non-autoregressive machine translation,'' in \emph{International Conference on Learning Representations}, 2019.

\bibitem{ren2020study}
Y.~Ren, J.~Liu, X.~Tan, Z.~Zhao, S.~Zhao, and T.-Y. Liu, ``A study of non-autoregressive model for sequence generation,'' in \emph{Annual Meeting of the Association for Computational Linguistics}, 2020.

\bibitem{stern2019insertion}
M.~Stern, W.~Chan, J.~Kiros, and J.~Uszkoreit, ``Insertion transformer: Flexible sequence generation via insertion operations,'' in \emph{International Conference on Machine Learning}, 2019.

\bibitem{gu2019levenshtein}
J.~Gu, C.~Wang, and J.~Zhao, ``Levenshtein transformer,'' in \emph{Advances in Neural Information Processing Systems}, 2019.

\bibitem{ghazvininejad2019mask}
M.~Ghazvininejad, O.~Levy, Y.~Liu, and L.~Zettlemoyer, ``Mask-predict: Parallel decoding of conditional masked language models,'' in \emph{Proceedings of the 2019 Conference on Empirical Methods in Natural Language Processing and the 9th International Joint Conference on Natural Language Processing (EMNLP-IJCNLP)}.\hskip 1em plus 0.5em minus 0.4em\relax Association for Computational Linguistics, 2019.

\bibitem{wang2018semi}
C.~Wang, J.~Zhang, and H.~Chen, ``Semi-autoregressive neural machine translation,'' in \emph{Conference on Empirical Methods in Natural Language Processing}, 2018.

\bibitem{shu2020latent}
R.~Shu, J.~Lee, H.~Nakayama, and K.~Cho, ``Latent-variable non-autoregressive neural machine translation with deterministic inference using a delta posterior,'' in \emph{AAAI Conference on Artificial Intelligence}, 2020.

\bibitem{bao2022glat}
Y.~Bao, H.~Zhou, S.~Huang, D.~Wang, L.~Qian, X.~Dai, J.~Chen, and L.~Li, ``Glat: Glancing at latent variables for parallel text generation,'' in \emph{Annual Meeting of the Association for Computational Linguistics}, 2022.

\bibitem{ran2021guiding}
Q.~Ran, Y.~Lin, P.~Li, and J.~Zhou, ``Guiding non-autoregressive neural machine translation decoding with reordering information,'' in \emph{AAAI Conference on Artificial Intelligence}, 2021.

\bibitem{bao2021non}
Y.~Bao, S.~Huang, T.~Xiao, D.~Wang, X.~Dai, and J.~Chen, ``Non-autoregressive translation by learning target categorical codes,'' in \emph{Annual Conference of the Nations of the Americas Chapter of the Association for Computational Linguistics}, 2021.

\bibitem{xiao2022survey}
Y.~Xiao, L.~Wu, J.~Guo, J.~Li, M.~Zhang, T.~Qin, and T.-y. Liu, ``A survey on non-autoregressive generation for neural machine translation and beyond,'' \emph{arXiv preprint}, 2022.

\bibitem{carion2020end}
N.~Carion, F.~Massa, G.~Synnaeve, N.~Usunier, A.~Kirillov, and S.~Zagoruyko, ``End-to-end object detection with transformers,'' in \emph{European Conference on Computer Vision}, 2020.

\bibitem{wang2020axial}
H.~Wang, Y.~Zhu, B.~Green, H.~Adam, A.~Yuille, and L.-C. Chen, ``Axial-deeplab: Stand-alone axial-attention for panoptic segmentation,'' in \emph{European Conference on Computer Vision}, 2020.

\bibitem{zhu2020deformable}
X.~Zhu, W.~Su, L.~Lu, B.~Li, X.~Wang, and J.~Dai, ``Deformable detr: Deformable transformers for end-to-end object detection,'' \emph{arXiv preprint}, 2020.

\bibitem{wang2022anchor}
Y.~Wang, X.~Zhang, T.~Yang, and J.~Sun, ``Anchor detr: Query design for transformer-based detector,'' in \emph{AAAI Conference on Artificial Intelligence}, 2022.

\bibitem{bai2022transfusion}
X.~Bai, Z.~Hu, X.~Zhu, Q.~Huang, Y.~Chen, H.~Fu, and C.-L. Tai, ``Transfusion: Robust lidar-camera fusion for 3d object detection with transformers,'' in \emph{IEEE Conference on Computer Vision and Pattern Recognition}, 2022.

\bibitem{devlin2018bert}
J.~Devlin, M.-W. Chang, K.~Lee, and K.~Toutanova, ``Bert: Pre-training of deep bidirectional transformers for language understanding,'' \emph{arXiv preprint}, 2018.

\bibitem{he2016deep}
K.~He, X.~Zhang, S.~Ren, and J.~Sun, ``Deep residual learning for image recognition,'' in \emph{IEEE Conference on Computer Vision and Pattern Recognition}, 2016.

\bibitem{chang2019argoverse}
M.-F. Chang, J.~Lambert, P.~Sangkloy, J.~Singh, S.~Bak, A.~Hartnett, D.~Wang, P.~Carr, S.~Lucey, D.~Ramanan \emph{et~al.}, ``Argoverse: 3d tracking and forecasting with rich maps,'' in \emph{IEEE Conference on Computer Vision and Pattern Recognition}, 2019.

\bibitem{lee2020centermask}
Y.~Lee and J.~Park, ``Centermask: Real-time anchor-free instance segmentation,'' in \emph{IEEE Conference on Computer Vision and Pattern Recognition}, 2020.

\bibitem{chen2018encoder}
L.-C. Chen, Y.~Zhu, G.~Papandreou, F.~Schroff, and H.~Adam, ``Encoder-decoder with atrous separable convolution for semantic image segmentation,'' in \emph{European Conference on Computer Vision}, 2018, pp. 801--818.

\bibitem{cordts2016cityscapes}
M.~Cordts, M.~Omran, S.~Ramos, T.~Rehfeld, M.~Enzweiler, R.~Benenson, U.~Franke, S.~Roth, and B.~Schiele, ``The cityscapes dataset for semantic urban scene understanding,'' in \emph{IEEE Conference on Computer Vision and Pattern Recognition}, 2016, pp. 3213--3223.

\bibitem{wang2021fcos3d}
T.~Wang, X.~Zhu, J.~Pang, and D.~Lin, ``Fcos3d: Fully convolutional one-stage monocular 3d object detection,'' in \emph{IEEE International Conference on Computer Vision}, 2021.

\bibitem{ko2021key}
Y.~Ko, Y.~Lee, S.~Azam, F.~Munir, M.~Jeon, and W.~Pedrycz, ``Key points estimation and point instance segmentation approach for lane detection,'' \emph{IEEE Transactions on Intelligent Transportation Systems}, vol.~23, no.~7, pp. 8949--8958, 2021.

\bibitem{russakovsky2015imagenet}
O.~Russakovsky, J.~Deng, H.~Su, J.~Krause, S.~Satheesh, S.~Ma, Z.~Huang, A.~Karpathy, A.~Khosla, M.~Bernstein \emph{et~al.}, ``Imagenet large scale visual recognition challenge,'' \emph{International Journal of Computer Vision}, 2015.

\end{thebibliography}
